\documentclass{midl} % Include author names

\usepackage{mwe} % to get dummy images
\usepackage[table]{xcolor}
\usepackage{xcolor}
\usepackage{longtable}
\usepackage{multirow}
\usepackage{dirtytalk}
\usepackage{booktabs}
\usepackage{float}
\usepackage{pifont}
\usepackage{amssymb}
\usepackage{graphicx}
\usepackage[dvipsnames]{xcolor}
\usepackage{mdframed}
\usepackage{wrapfig}
\usepackage{soul}
\definecolor{grey}{rgb}{0.5,0.5,0.5}
\newcommand{\cmark}{\ding{51}}% check mark
\newcommand{\xmark}{\ding{55}}% cross mark

\jmlryear{2026}
\jmlrworkshop{Full Paper -- MIDL 2026}
\jmlrvolume{-- 280}\editors{Accepted for publication at MIDL 2026}

\title[RadImageNet-VQA]{RadImageNet-VQA: A Large-Scale CT and MRI Dataset for Radiologic Visual Question Answering}

\midlauthor{
\Name{Léo Butsanets\nametag{$^{1}$}} \Email{leo.butsanets@raidium.eu} \\
\Name{Charles Corbière\nametag{$^{1}$}} \Email{charles.corbiere@raidium.eu} 
\\
\Name{Julien Khlaut\nametag{$^{1,2,3}$}} \Email{julien.khlaut@raidium.eu} 
\\
\Name{Pierre Manceron\nametag{$^{1}$}} \Email{pierre.manceron@raidium.eu}
\\
\Name{Corentin Dancette\nametag{$^{1}$}} \Email{corentin.dancette@raidium.eu} 
\\
\addr $^{1}$Raidium, Paris, France. \\
\addr $^{2}$Université de Paris Cité, Hôpital Européen Georges Pompidou, AP-HP, Paris, France. \\
\addr $^{3}$Department of Vascular and Oncological Interventional Radiology, INSERM, Paris, France.
}

\begin{document}

\maketitle

\begin{abstract}

In this work, we introduce \textit{RadImageNet-VQA}, a large-scale dataset designed to advance radiologic visual question answering (VQA) on CT and MRI exams. While existing medical VQA datasets are limited in scale, dominated by X-ray imaging or biomedical illustrations, and prone to text-based shortcuts, RadImageNet-VQA is built from expert-curated annotations and provides 750K images paired with 7.5M QA samples. It covers three key tasks—abnormality detection, anatomy recognition, and pathology identification—spanning 8 anatomical regions and 97 pathology categories, and supports open-ended, closed-ended, and multiple-choice questions. Extensive experiments show that state-of-the-art vision-language models still struggle with fine-grained pathology identification, especially in open-ended settings and even after fine-tuning. Text-only analysis further reveals that model accuracies collapse to near-random without image inputs, confirming that RadImageNet-VQA is free from linguistic shortcuts. The full dataset and benchmark are publicly available at \url{https://huggingface.co/datasets/raidium/RadImageNet-VQA}.

\end{abstract}

\begin{keywords}
Medical Visual Question Answering, Vision-Language Models, Radiology.
\end{keywords}

%==================================================================
\section{Introduction}
Interpreting radiology exams requires integrating complex visual patterns with clinical reasoning, often with medical context~\cite{multimodalhealthcare2024,dancette2025curiamultimodalfoundationmodel,khlaut2025radsam}.
Vision-language models (VLMs)~\cite{blip2,bai2025qwen2,chen2024internvl,llava} have recently achieved strong performance on general multimodal benchmarks. Medical variants, such as MedGemma~\cite{sellergren2025medgemma} and Lingshu~\cite{xu2025lingshu}, now incorporate curated medical data to better align with clinical knowledge. These models are particularly attractive for radiology, where generating reports and supporting clinical reasoning require combining visual and textual information. Yet, evaluating free-text radiology reports remains difficult: commonly used similarity metrics often fail to capture clinical correctness, factual consistency, or semantic alignment with ground-truth findings~\cite{irvin2019chexpert, delbrouck2024radgraph, xu2025radeval}. In contrast, radiologic visual question answering (VQA) offers a more structured alternative to probe model reasoning. Its question-answer format mirrors how radiologists interrogate imaging findings, providing an interpretable way to evaluate a model’s image-grounded reasoning.

Existing medical VQA datasets~\cite{vqarad, liu2021slake,pal2025rexvqa,zhang2023pmc} present several limitations that hinder proper evaluation of these models. Early efforts rely on limited expert annotation and thus offer small collections with narrow anatomical and pathological coverage~\cite{vqarad, liu2021slake}. Larger recent datasets draw heavily from X-ray images or biomedical figures scraped from publications~\cite{pal2025rexvqa, zhang2023pmc}, providing little representation of other common imaging modalities, in particular CT or MRI. Many also contain textual shortcuts that allow models to answer correctly without interpreting the image, as shown in Section~\ref{subsection:text-shortcut}. Beyond evaluation, progress in medical VLMs also requires access to large, well-curated training data aligned with diagnostic tasks. This emphasizes the need for large-scale VQA resources that spans multiple radiologic tasks while minimizing possible shortcut cues to offer a more rigorous foundation for assessing and advancing medical VLMs.

\begin{figure}
    \centering      
    \includegraphics[width=1\linewidth]{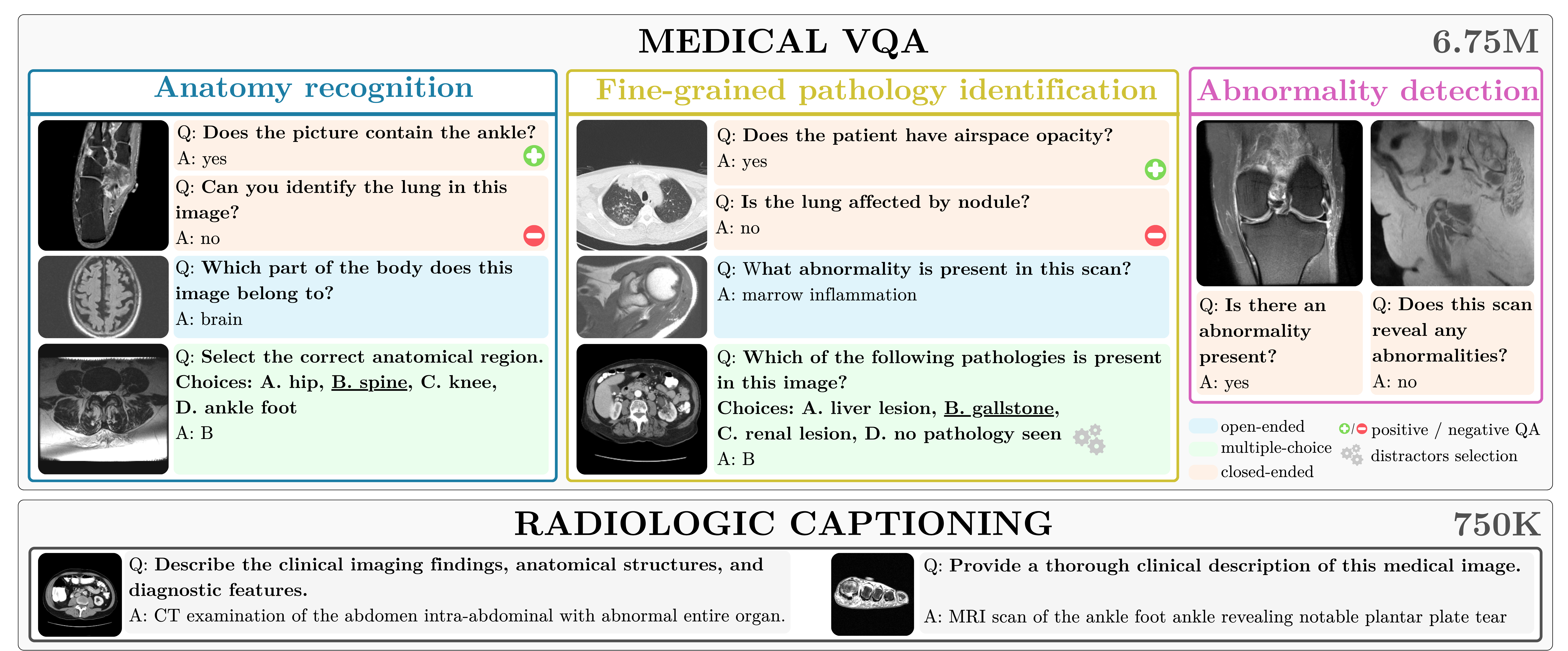}
    \vspace{-0.7cm}
    \caption{\textbf{Overview of the RadImageNet-VQA dataset}, which provides radiology-focused supervision across three VQA tasks (anatomy recognition, fine-grained pathology identification, and abnormality detection) using diverse open-ended, closed-ended, and multiple-choice formats.  It also includes radiologic captioning pairs for image–text alignment.
    }
    \vspace{-0.7cm}
    \label{fig:introduction}
\end{figure}

In this paper, we introduce \emph{RadImageNet-VQA}, a large-scale dataset designed for training and benchmarking radiologic VQA on CT and MRI exams (see Figure~\ref{fig:introduction}). Built from the CT/MRI subset of RadImageNet~\cite{mei2022radimagenet} and its expert-curated anatomical and pathological annotations, RadImageNet-VQA provides 750K images with 7.5M generated samples, including 750K medical captions for visual-text alignment and 6.75M question-answer pairs that span three radiology tasks: fine-grained pathology identification, anatomy recognition, and abnormality detection. The dataset includes open-ended, closed-ended, and multiple-choice questions across 8 anatomical regions and 97 pathologies, where questions were generated with prompt-based templates and constructed to probe visual-grounded understanding while minimizing text-only shortcut answering. For evaluation, we construct a stratified benchmark of 1,000 images with 9,000 question-answer pairs covering all tasks and question types. Through extensive zero-shot evaluation and fine-tuning experiments with RadImageNet-VQA, we show that while anatomy recognition is nearly solved for current VLMs, pathology identification remains a major bottleneck, especially for open-ended responses. Text-only analysis further confirm that RadImageNet-VQA greatly reduces shortcut cues present in existing datasets. Fine-tuning yields substantial gains across models, yet we observe that initializing models with medically pretrained vision encoders such as MedSigLIP~\cite{sellergren2025medgemma} provides no advantage over general-purpose encoders.

Our main contributions are as follows:

\vspace{0.1em}
$\bullet$ We release RadImageNet-VQA, a radiologic dataset that includes both a curated benchmark and a large training corpus. It spans 8 anatomical regions, 97 pathologies, and three task families (abnormality, anatomy, pathology) with multiple question formats, providing 750K images paired with 7.5M generated samples.

\vspace{0.1em}
$\bullet$ We conduct an extensive evaluation of state-of-the-art VLMs showing that existing models perform well on anatomy and basic abnormality detection but still struggle with fine-grained pathology identification. Text-only ablations confirm that RadImageNet-VQA minimizes shortcut cues.

\vspace{0.1em}
$\bullet$ Fine-tuning on RadImageNet-VQA produces substantial gains across all model families, while medical pretrained vision encoders do not improve downstream performance.

%==================================================================
\section{Related Work}

%------------------------------------------------------------------
\subsection{Vision-language models in radiology}

Recent VLMs, such as LLaVA~\cite{llava}, QwenVL~\cite{bai2025qwen2}, InternVL~\cite{chen2024internvl}, typically combine a vision encoder, often initialized from CLIP~\cite{radford2021learning} or SigLIP~\cite{zhai2023sigmoid} backbones, with a large language model (LLM) via a lightweight adapter module. In the medical domain, progress has largely been driven by data-centric adaptation, where general-purpose VLMs are aligned to medical images through image–text pretraining and multimodal instruction tuning~\cite{llavamed, sellergren2025medgemma, xu2025lingshu}. While several 3D radiology VLMs have recently been proposed~\cite{blankemeier2024merlin, radfm, xin2025med3dvlm, ates2025dcformer}, their development is constrained by the scarcity of large volumetric datasets, making broad and diverse 2D datasets still essential for training and evaluation. Beyond training, reliable evaluation also remains challenging. Radiology report generation~\cite{bannur2024maira2groundedradiologyreport} is appealing but difficult to assess automatically, as standard metrics correlate poorly with clinical correctness. On the other hand, visual-question answering (VQA) provides a more controlled and interpretable proxy framework for probing clinical reasoning grounded in images. This has motivated increasing interest in radiologic VQA benchmarks, particularly for CT and MRI, which remain underrepresented in existing resources.

\begin{table}[t]
\centering
\scriptsize
\renewcommand{\arraystretch}{1.3}

\resizebox{\linewidth}{!}{%
\begin{tabular}{l l l r r r}
\toprule
\textbf{Dataset} &
\textbf{Modality} &
\textbf{Question Types} &
\textbf{\# Images} &
\textbf{\# QAs Train} &
\textbf{\# QAs Test} \\
\midrule

VQA-RAD~\cite{vqarad} &
X-ray, CT &
open, closed &
0.3K &
2.8K &
0.45K \\

SLAKE~\cite{liu2021slake} &
X-ray, CT, MRI &
open, closed &
0.6K &
4.9K &
1.1K \\

MMMU-Med~\cite{yue2024mmmu} &
diverse &
multiple-choice &
1.9K &
-- &
1.8K \\

ReXVQA~\cite{pal2025rexvqa} &
X-ray &
multiple-choice &
160K &
573K &
40.8K \\

M3D-VQA~\cite{m3d} &
3D CT &
multiple-choice  &
96K  &
662K &
2K \\

3D-RAD~\cite{3drad} &
3D CT  &
open, closed  &
16K  &
136K &
34K \\

\rowcolor{gray!10}
\textbf{RadImageNet-VQA (Ours)} &
\textbf{CT, MRI} &
\textbf{\parbox[t]{2.5cm}{open, closed,\\multiple-choice}} &
\textbf{750K} &
\textbf{7.5M} &
\textbf{9K} \\
\bottomrule
\end{tabular}%
}
\vspace{-0.3cm}
\caption{\textbf{Comparison of radiologic VQA datasets}
compared by imaging modality, supported question types, and number of images, and the number of question-answer pairs for training and evaluation. RadImageNet-VQA provides the largest CT/MRI coverage and the most extensive QA corpus.}
\vspace{-0.6cm}
\label{tab:datasets}
\end{table}

\subsection{Medical VQA datasets}
Early radiologic VQA, such as VQA-RAD~\cite{vqarad} and SLAKE~\cite{liu2021slake} introduced structured QA annotations but are limited by scale and by narrow anatomical and pathological coverage. More recent efforts, such as the medical subset of MMMU~\cite{yue2024mmmu} extend medical breadth, yet their radiology content is sparse and relies largely on images from textbooks or publications rather than CT or MRI. Programmatic datasets such as ReXVQA~\cite{pal2025rexvqa} improve scalability but are restricted to X-ray images and therefore cover only a small portion of radiology practice. Table~\ref{tab:datasets} summarizes these datasets and highlights the narrow CT/MRI coverage of existing benchmarks. Additionally, a persistent issue across these benchmarks is the presence of linguistic shortcuts, label phrasing, or distributional biases that allow models to guess without relying on images
In parallel, several 3D medical VQA benchmarks, such as DeepTumorVQA~\cite{deeptumor}, M3D-VQA~\cite{m3d}, and 3D-RAD~\cite{3drad}, extend to volumetric tasks. Although clinically meaningful, these datasets cover limited anatomy and pathologies, require extensive expert curation, and remain small in scale. Current 3D VLMs are trained on small volumetric corpora and lack the broad visual priors available to large 2D models. Strengthening 2D CT/MRI supervision therefore remains essential for building radiology VLMs, even as 3D approaches continue to mature.

%==================================================================
\section{RadImageNet-VQA Dataset}

In this section, we describe the construction process of RadImageNet-VQA dataset, as illustrated in Figure~\ref{fig:RadImageNet-VQA}, and detail the statistics and metrics of the benchmark subset.

%------------------------------------------------------------------
\subsection{Dataset construction pipeline}

Data is sourced from RadImageNet~\cite{mei2022radimagenet}, a large expert-annotated medical imaging dataset in which each image is associated with a modality (CT, MRI, US), a body part (e.g., abdomen, hip, brain) and a pathology label. From this resource, we use the CT and MRI subsets to form the basis for generating clinically meaningful captions and VQA samples across anatomy, abnormality, and fine-grained pathology tasks. \\

\begin{figure}[t]
    \centering
        \vspace{-0.7cm}
    \includegraphics[width=1\linewidth]{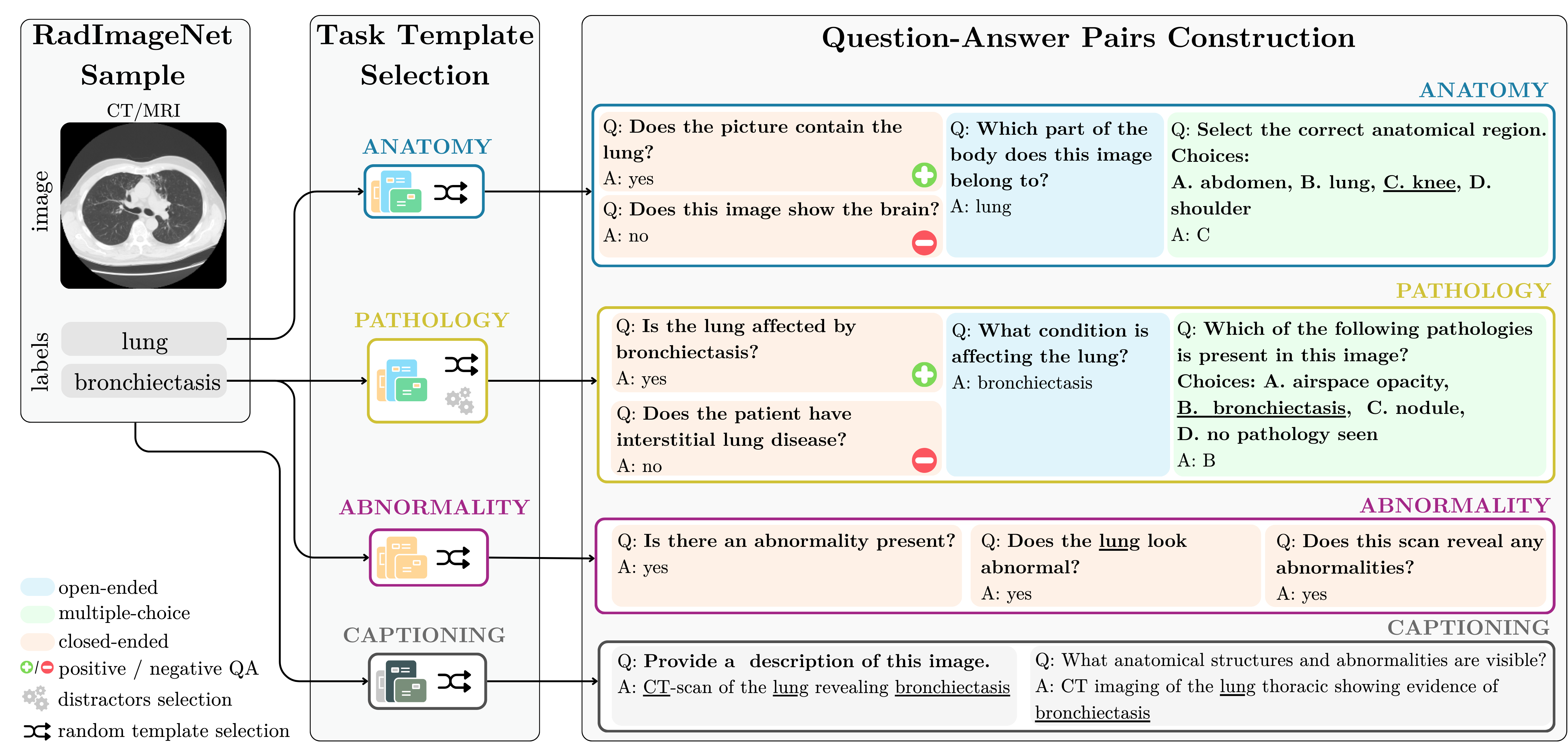}
    \vspace{-0.7cm}
   \caption{
    \textbf{RadImageNet-VQA construction pipeline.} Expert-annotated CT/MRI images are converted into radiology-aware captions and VQA samples using task- and format-specific templates. The pipeline generates open-ended, closed-ended, and multiple-choice questions across anatomy recognition, abnormality detection, and pathology identification, with distractors designed to prevent shortcuts.
    } 
    \vspace{-0.4cm}
    \label{fig:RadImageNet-VQA}
\end{figure}

\noindent
\textbf{Radiologic captioning generation.}
VLMs are typically trained with an alignment stage in which the model learns to associate visual content with textual semantics using large collections of image–caption pairs. As presented in Figure \ref{fig:RadImageNet-VQA}, to support a clinically meaningful alignment dataset in training, we leverage the rich metadata in RadImageNet, including acquisition modality, anatomical region, and pathology category, and convert it into structured radiologic captions. Each image is paired with a synthesized textual description generated from these metadata fields sampling randomly from a diverse set of radiology-aware templates. For example, modality and anatomy tags are verbalized in formulations like \texttt{"A [modality] scan of the [anatomy] showing [pathology]”}. \\

\noindent
\textbf{VQA sample generation.}
To support both instruction tuning and benchmark construction, each annotated image is converted into structured VQA samples using task-specific question-answer templates. As illustrated in Figure~\ref{fig:RadImageNet-VQA}, we generate closed-ended, open-ended, and multiple-choice questions across three complementary tasks: \textit{anatomy recognition}, \textit{abnormality detection} and \textit{pathology identification}. Anatomy recognition identifies the imaged region, abnormality detection determines whether an image contains any abnormal finding, and pathology identification—by far the most challenging—requires distinguishing between specific diseases within a given anatomical context.

For each task, we design a set of template-based question formulations for each question type (Appendix~\ref{appx_subsec:question_templates}). For every task and question type, we define between 2 and 7 linguistic variations, ensuring diversity while reducing opportunities for models to exploit textual shortcuts. For closed-ended anatomy and pathology questions, we generate both positive and negative variants, where the expected answer is respectively “yes” (the organ or pathology is present) or “no” (it is absent). These paired formulations allow us to probe VLMs more precisely, particularly with respect to biases such as hallucinating organs or pathologies even when they are not present.

\begin{figure}[t]
    \centering
    \includegraphics[width=0.99
\linewidth]{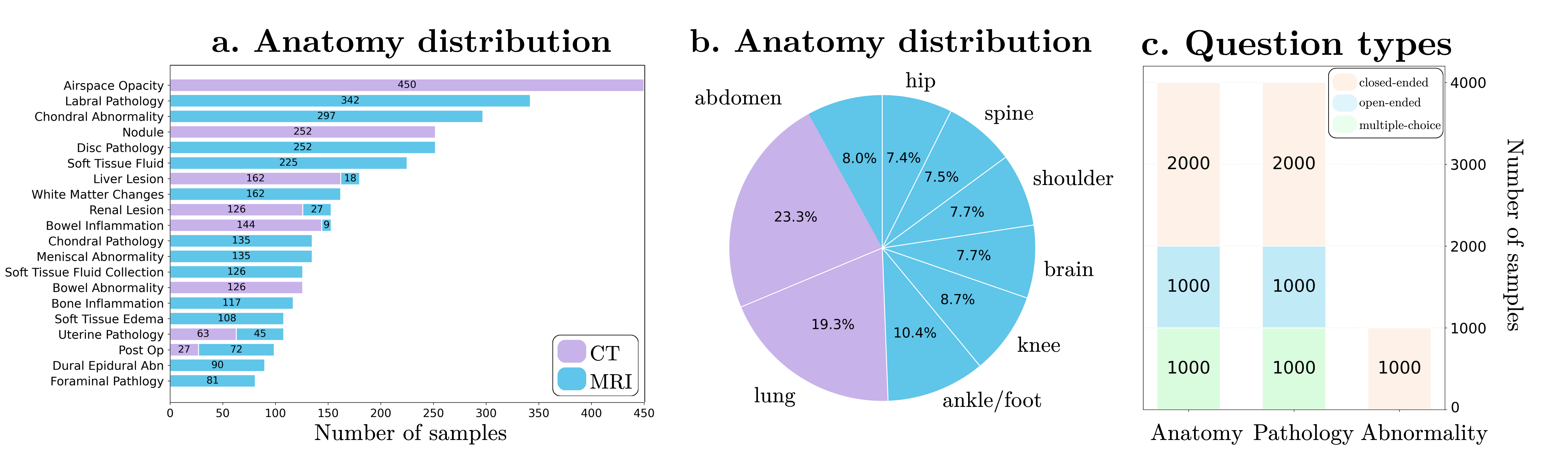}
    \vspace{-0.5cm}
    \caption{\textbf{Composition of the RadImageNet-VQA benchmark}, containing 1,000 CT/MRI images and 9,000 QA pairs: 
(a) most frequent pathology labels, 
(b) anatomy distribution,
(c) question-type distribution.
}
    \vspace{-0.6cm}
    \label{fig:bench_distrib}
\end{figure}

Multiple-choice questions require careful construction as distractors, \textit{i.e.} incorrect answer options, must be chosen in a way that prevents trivial elimination. For anatomy recognition, distractors are sampled from other existing anatomical regions in the dataset. For pathology identification, distractors are restricted to clinically plausible diseases from the same region, ensuring that the model cannot solve the question by matching anatomy alone. We also include the option \emph{“no pathology seen”} in every pathology multiple-choice. As shown in Appendix~\ref{appx_subsec:add-distractors}, increasing the frequency of this option in multiple-choice options causes a clear and consistent drop in accuracy for medical-oriented models (e.g., MedGemma and Lingshu), while general-purpose models remain comparatively stable. This suggests that medical VLMs often assume that pathology is present and treat “no pathology” as unlikely instead of confirming it visually. %When such a distractor appears frequently in abnormal cases, that shortcut is reduced, revealing weaker visual grounding. 
Including this option therefore reduces this bias and encourages models to distinguish true pathology from its absence, resulting in a more challenging and faithful evaluation. \\

\noindent
\textbf{Training set creation.}
We apply the captioning and VQA generation pipeline to the full RadImageNet
training split to construct a large-scale corpus for multimodal model training.
This yields structured radiologic captions for medical visual-text alignment and extensive VQA data for instruction tuning. In total, the training corpus comprises approximately 750K images with 7.5M samples (750K image-captions and 6.75M QA pairs). A complete list of anatomical regions and pathology categories is provided in Appendix~\ref{appx_subsec:detailed_anatomy_pathologies}.

\begin{table}[t]
\centering
\scriptsize
\setlength{\tabcolsep}{4pt} 
\renewcommand{\arraystretch}{1.2}
\begin{tabular}{l|cccc|c|cccc|c}
\toprule
\textbf{Model} 
& \multicolumn{4}{c|}{\textbf{Anatomy}} 
& \textbf{Abn}
& \multicolumn{4}{c|}{\textbf{Pathology}} 
& \textbf{Avg} \\
\cmidrule(lr){2-5} \cmidrule(lr){6-6} \cmidrule(lr){7-10}
& Open & Closed+ & Closed– & MC
& Closed
& Open & Closed+ & Closed– & MC
& \\
\midrule
\rowcolor{gray!10}
\multicolumn{11}{l}{\textbf{General-purpose models}} \\
\rowcolor{lightgray!5}
LLaVA-OneVision-Qwen2-7B
& 48.4 & 82.7 & 81.3 & 88.7
& 49.8
& 16.0 & 55.3 & 61.3 & 33.6
& 57.5 \\

Qwen2.5-VL-3B-Instruct
& 37.7 & 83.7 & 77.1 & 77.9
& 70.5
& 10.0 & 78.1 & 21.4 & 34.8
& 54.6 \\

Qwen2.5-VL-7B-Instruct
& 37.5 & 84.9 & 79.1 & 80.5
& 69.5
&  9.8 & 69.2 & 47.4 & 30.1
& 56.4 \\

InternVL3.5-8B
& 50.9 & \underline{98.1} & 75.9 & \textbf{93.3}
& 58.9
&  9.9 & \underline{85.9} & 27.8 & 41.8
& 60.3 \\

InternVL3.5-14B
& 56.6 & \textbf{98.2} & 74.4 & 89.9
& \textbf{74.4}
& 11.7 & \textbf{86.7} & 33.7 & \textbf{47.1}
& \textbf{63.6} \\

GPT-5
& 44.3 & 72.4 & 81.8 & 89.3
& 27.5
& 15.8 & 54.9 & 68.3 & 41.2
& 54.9 \\

Gemini 2.5 Pro
& \textbf{65.7} & 76.5 & 81.9 & 88.8
& 17.8
& \underline{21.1} & 50.2 & 30.1 & 44.4
& 52.9 \\

\midrule
\rowcolor{gray!10}
\multicolumn{11}{l}{\textbf{Medical-oriented models}} \\

LLaVA-Med-v1.5-mistral-7b
& 44.3 & 89.9 & 55.3 & 58.1
& 22.4
& 10.2 & 41.8 & 66.6 & 26.4
& 48.2 \\

HuatuoGPT-Vision-7B
& 45.4 & 82.5 & \underline{89.0} & 88.3
& 60.6
& 13.6 & 65.5 & 69.2 & \underline{44.6}
& 48.9 \\

medgemma-4b-it
& \underline{62.9} & 76.4 & 82.5 & 84.8
& 55.4
& \textbf{30.6} & 54.2 & 77.4 & 36.8
& 51.5 \\

Lingshu-7B
& 49.6 & 90.7 & 85.1 & 88.9
& 47.9
& 15.7 & 57.0 & \underline{78.8} & 29.6
& \underline{60.4} \\

Lingshu-32B
& 45.2 & 75.5 & \textbf{92.1} & \underline{89.3}
& 54.5
& 14.4 & 46.4 & \textbf{88.8} & 31.7
& 59.8 \\

\bottomrule

\end{tabular}
\vspace{-0.3cm}
\caption{
\textbf{Zero-shot accuracies (\%) of VLMs on RadImageNet-VQA benchmark.} 
Results are reported across anatomy recognition, abnormality detection (\textit{Abn}), and pathology identification using four question formats: 
\textit{Open} (free-form), 
\textit{Closed+} (always `yes’ as true answer) , 
\textit{Closed–} (always `no’), 
and \textit{MC} (multiple-choice).}
\vspace{-0.6cm}
\label{tab:zero-shot}
\end{table}

%------------------------------------------------------------------
\subsection{Benchmark}

For evaluation, we sample 1,000 CT/MRI images from the RadImageNet test split while preserving the distribution of anatomical regions, pathology categories, and abnormal cases. Applying our pipeline yields 9,000 closed-ended, open-ended, and multiple-choice QA pairs. Figure~\ref{fig:bench_distrib} shows the distribution of pathology labels, question formats, anatomical regions, and normal/abnormal cases. Samples are evenly distributed across tasks and question types, and the dataset includes both normal (28.1\%) and abnormal (71.9\%) studies across multiple anatomical regions. The most common pathology categories include a wide range of soft-tissue, musculoskeletal, and neuro-abdominal findings represented in CT and MRI.
\\

\paragraph{Evaluation metrics.}
Model predictions are scored as correct or incorrect, and mean accuracy is reported. Closed-ended (yes/no) questions are judged by checking for the expected token. Open-ended responses are evaluated using \textit{LLM-as-a-judge}~\cite{llmjudge2023} framework against ground-truth (Appendix~\ref{appx_subsec:llm-judge-prompt}). 
For multiple-choice questions, we extract the predicted option letter with rule-based parser and compare it to the true answer.

To assess the reliability of the LLM judge on open-ended samples, we conducted a human validation on a stratified subset of 380 examples, balanced across the 10 evaluated models and LLM judge outcomes. Each example (question, ground-truth answer, and model response) correctness was independently annotated by two reviewers, blinded to both model identity and LLM judge decisions. The LLM judge matches human evaluation in 90.6\% of cases, achieving a substantial agreement with a Cohen’s $\kappa$ score of 0.72, supporting its use for open-ended evaluation in the benchmark. Validation details are provided in Appendix~\ref{appx_subsec:llm_judge_validation}.
%==================================================================
\section{Experiments}

We evaluate state-of-the-art VLMs on RadImageNet-VQA under zero-shot and fine-tuned settings, and analyze text shortcuts, data-sampling strategies, and vision encoder choices.

%------------------------------------------------------------------
\subsection{Zero-shot evaluation}

We benchmark a wide range of general-purpose and medical-oriented VLMs, including open-source models (e.g., LLaVA-OneVision, Qwen2.5-VL, InternVL) and proprietary systems (e.g., GPT-5, Gemini 2.5 Pro) on the full RadImageNet-VQA benchmark. Models generate free-text answers for open-ended questions and letter choices for multiple-choice questions. Accuracy is computed via exact match for closed and MC questions and LLM-as-a-judge scoring for open-ended responses. Results are displayed in Table~\ref{tab:zero-shot}.

First, pathology identification emerges as the primary bottleneck. While models perform well on anatomy recognition (e.g., InternVL3.5-8B~\cite{chen2024internvl} achieves 93.3\% on anatomy multiple-choice), they struggle severely with fine-grained disease identification. Open-ended pathology questions prove particularly challenging, with most models scoring below 20\% accuracy, and even the best performer on this task, MedGemma-4b~\cite{sellergren2025medgemma}, attains only 30.6\%. These results underscore that discriminating between specific pathologies based on visual features remains a major unsolved problem.

Then, we observe a nuanced performance gap between general-purpose and medical-oriented models. General-purpose models achieve the highest average accuracy with InternVL3.5 -14B (63.6\%) and lead in several categories, including abnormality detection and closed-ended pathology questions. Medical-oriented models outperform general ones only in some specific areas, such as MedGemma's superior performance on open-ended pathology questions. This suggests that scale and broad visual–language pretraining still provide a stronger foundation than existing medical specialization strategies.

Finally, the results reveal that even the most capable general-purpose models lack essential radiologic skills. GPT-5 (27.5\%) and Gemini 2.5 Pro (17.8\%) perform poorly on abnormality detection, below random guessing. We intuit that aggressive safety alignment may suppress the model's willingness to identify abnormalities, causing them to default to conservative “no abnormality” responses rather than rely on visual evidence.

%------------------------------------------------------------------
\subsection{Text shortcuts analysis}
\label{subsection:text-shortcut}

\begin{figure}[b]
    \centering
    \includegraphics[width=0.98\linewidth]{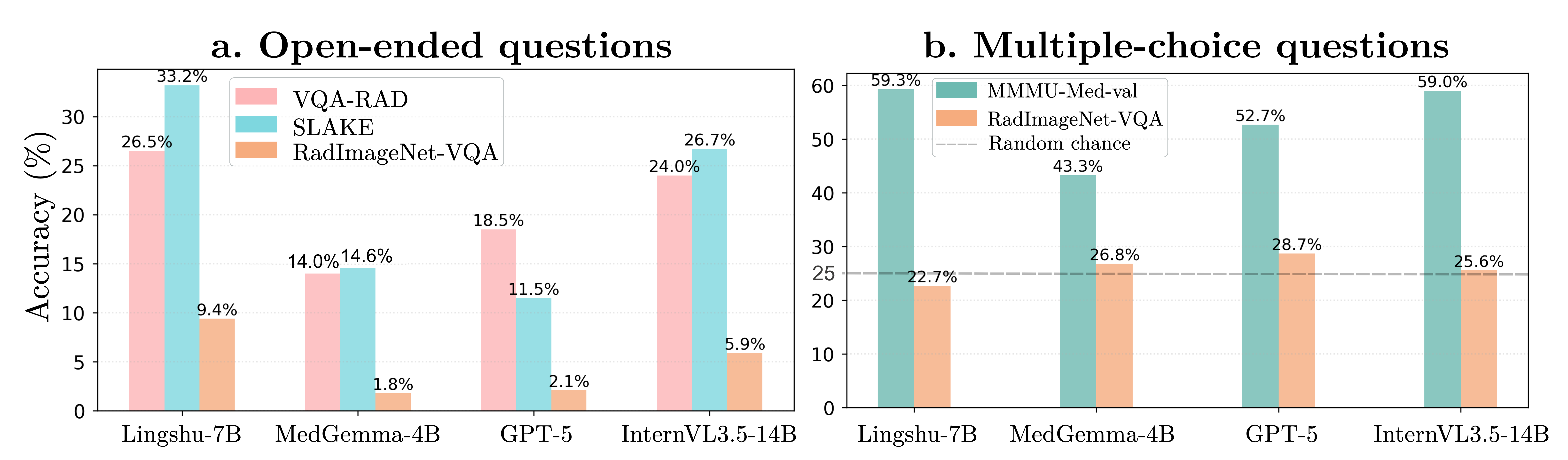}
    \vspace{-0.5cm}
   \caption{\textbf{Text-only analysis of multiple VLMs' accuracy} for open-ended and MC questions on RadImageNet-VQA, VQA-RAD, SLAKE, and MMU-Med-val.}
   \vspace{-0.5cm}
   \label{fig:text-ony-open-mc}
\end{figure}

To assess whether RadImageNet-VQA can be solved from linguistic priors alone, we evaluate several VLMs in a text-only setting, where images are removed and models receive only the question. 
We compare performance on VQA-RAD~\cite{vqarad}, SLAKE~\cite{liu2021slake}, MMMU-Med-val~\cite{yue2024mmmu}, and RadImageNet-VQA benchmarks to quantify the extent to which different datasets allow text-based shortcutting. (see Figure~\ref{fig:text-ony-open-mc})

On VQA-RAD and SLAKE, models reach 11.5–33\% text-only accuracy, indicating that many answers can be partially inferred from linguistic regularities. By contrast, open-ended accuracy on RadImageNet-VQA drops to near-random levels (2–10\%) across all models, confirming that shortcut cues are largely removed. The text-only evaluation also reveals differences in model behavior. Some models, e.g. Lingshu-7B and InternVL variants, maintain non-zero accuracy on RadImageNet-VQA, but qualitative inspection shows that these predictions reflect dataset priors rather than uncertainty awareness. Models rarely acknowledge missing visual information and instead commit to plausible guesses based on high-frequency anatomical regions or semantic variants of pathology terms, rather than image-grounded reasoning. Some examples are presented in Appendix~\ref{appx_subsec:examples}.

The same pattern holds for multiple-choice questions. On MMMU-Med-val, text-only accuracy remains noticeably above random, reflecting reliance on textual priors or weak distractor design. In contrast, RadImageNet-VQA collapses to the expected 25\% baseline, indicating that models cannot exploit option distributions or question phrasing. We can further observe in Appendix~\ref{appx_fig:text-only-mc-abn-vs-norm} that residual correct answers are largely driven by conservative defaults—such as selecting “no pathology seen” when no image is available—which yield correct predictions on normal cases but fail on abnormal ones. General-purpose models show a flatter spread across options, consistent with near-random guessing.

Overall, these findings show that RadImageNet-VQA substantially suppresses linguistic shortcuts in both open-ended and multiple-choice formats. When deprived of images, models fail to recover anatomical or pathological information and instead rely on dataset frequency or generic medical priors. This confirms that RadImageNet-VQA requires image-grounded interpretation rather than text-driven guessing.

%------------------------------------------------------------------
\subsection{Fine-tuning on RadImageNet-VQA}
\label{sec:model-fine-tuning}

%------------------------------------------------------------------
\subsubsection{Experimental setup}
\begin{wrapfigure}{r}{0.45\textwidth}
\vspace{-0.6cm}
  \begin{center}
    \includegraphics[width=\linewidth]{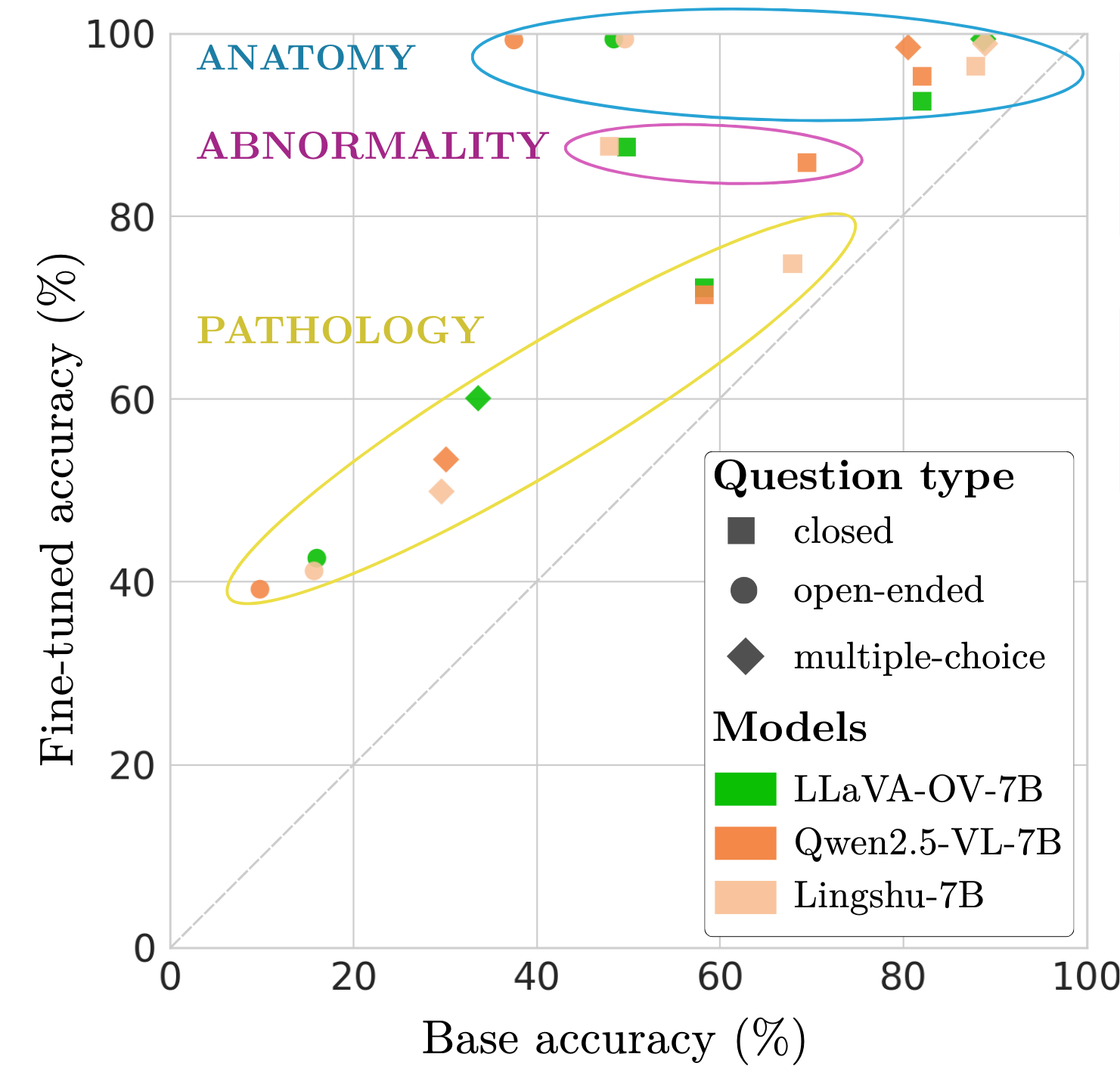}
  \end{center}
  \vspace{-0.8cm}
  \caption{Comparison of base and fine-tuned accuracies on RadImageNet-VQA.}
  \vspace{-0.6cm}
   \label{fig:ft-vs-base-merge-closed}
\end{wrapfigure}
As training data, we assemble a multimodal corpus combining:
(1) RadImageNet-VQA train split, (2) CT/MRI datasets converted into 2D VQA pairs-- KiTS22~\cite{kits23} and AbdomenAtlas~\cite{chen2025scaling})--
and (3) existing radiologic VQA datasets --VQA-RAD~\cite{vqarad}, SLAKE~\cite{liu2021slake}--, and the radiology subset of LLaVA-Med~\cite{llavamed}. Additional details and workflow are provided in Appendix~\ref{appx_sec:ft-details}.

We fine-tune two widely used open-source VLMs, LLaVA-OneVision~\cite{li2024llava} and Qwen2.5-VL-7B-Instruct~\cite{bai2025qwen2}, and a medical-oriented model, Lingshu-7B~\cite{xu2025lingshu}. All models employ of a SigLIP~\cite{zhai2023sigmoid} vision encoder that produces image embeddings. LLaVA-OneVision pairs it with a Qwen2-7B~\cite{yang2024qwen2technicalreport} language model
while Qwen2.5-VL-7B is based on the more recent Qwen2.5-7B language model.
Lingshu-7B share the same architecture as Qwen2.5-VL-7B but has been pretrained and fine-tuned on data including medical data. This allows us to apply the same training recipes across architectures while comparing general-purpose versus medically-aligned initialization.

We follow the standard two-phase training paradigm used in recent multimodal tuning consisting of (i) an \emph{alignment stage}, where the LLM is frozen and only the vision encoder and projection layers are updated to adapt to radiologic features; and (ii) \emph{instruction tuning}, where the full model is fine-tuned on supervised VQA data.

%------------------------------------------------------------------
\subsubsection{Evaluation of fine-tuned models}

Fine-tuning on the composed radiologic corpus yields consistent and substantial performance gains across all models. As shown in Figure~\ref{fig:ft-vs-base-merge-closed}, every point shifts above the diagonal, indicating reliable gains regardless of architecture, with average increases of +19.5\% to +22.5\% (Appendix~\ref{appx_tab:ft_values_deltas_colored}). Task-level results reveal that anatomy recognition is nearly saturated after fine-tuning (98.5–99.4\% on multiple-choice), abnormality detection shows the largest relative gains (+16.4–39.8\%, reaching 85.9 – 87.7\%), while pathology identification remains the primary bottleneck, clustering at lower accuracy even post-training. These trends hold across model families, demonstrating both the generality of our training strategy and the limits of class-level supervision for fine-grained disease identification. Finally, Lingshu-7B and Qwen2.5-VL-7B converge to nearly identical performance, indicating that radiologic supervision, and not prior medical pretraining, is the main driver of downstream capability.

%------------------------------------------------------------------
\subsubsection{Ablations on vision encoder and data sampling}

\begin{figure}[t]
    \centering
    \includegraphics[width=1\linewidth]{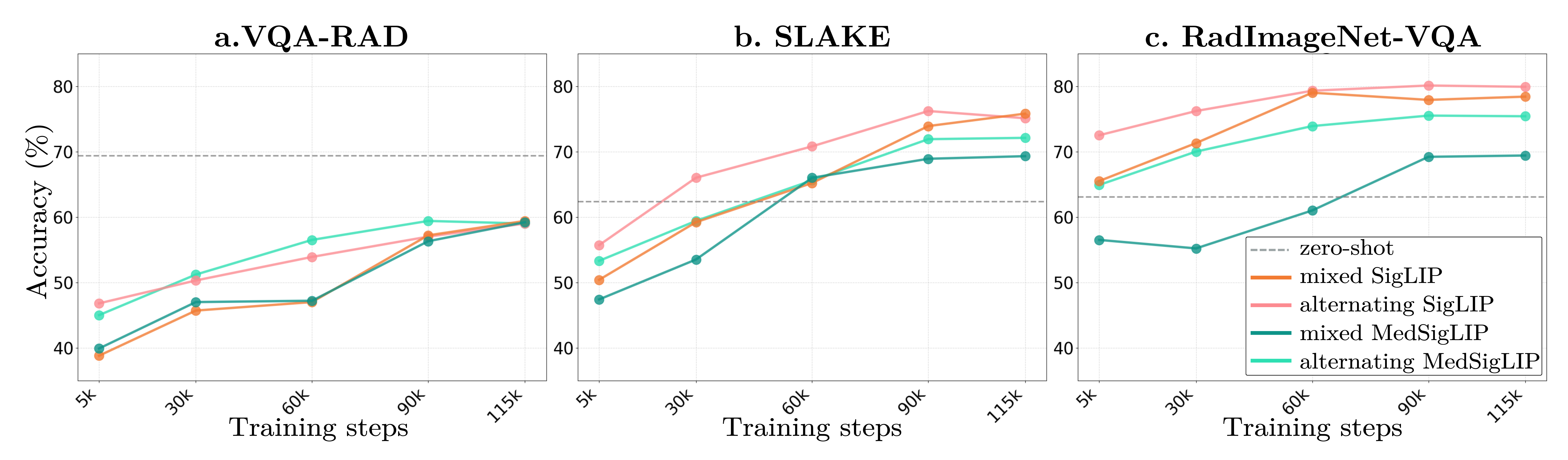}
    \vspace{-0.7cm}
    \caption{
\textbf{Impact of vision encoder and sampling with LLaVA-OneVision.} Alternating sampling with standard SigLIP yields the strongest performance.}
    \vspace{-0.5cm}
    \label{fig:ablation-ve-data}
\end{figure}

We compare four variants of LLaVA-OneVision variants using the standard SigLIP encoder or MedSigLIP~\cite{sellergren2025medgemma}, and with two different data-sampling strategies during fine-tuning. 
\textit{Mixed sampling} blends samples from all datasets within each batch, whereas \textit{alternating sampling} draws each batch from a single dataset and the source alternates across batches to balance exposure over time. Performance is measured across VQA-RAD, SLAKE, and RadImageNet-VQA and reported in Figure~\ref{fig:ablation-ve-data}. Models initialized with standard SigLIP outperform those using the medically pre-trained MedSigLIP, which may indicate that pretraining with multiple sources of medical images, including textbooks and X-ray, does not translate into systematic gains for VQA with CT and MRI. Alternating sampling data strategy yields slightly stronger and more stable results than mixed sampling, particularly early in training, likely because it reduces inter-dataset interference under small-batch constraints.

{\color{blue}

\begin{figure}[t]
    \centering
    \includegraphics[width=1\linewidth]{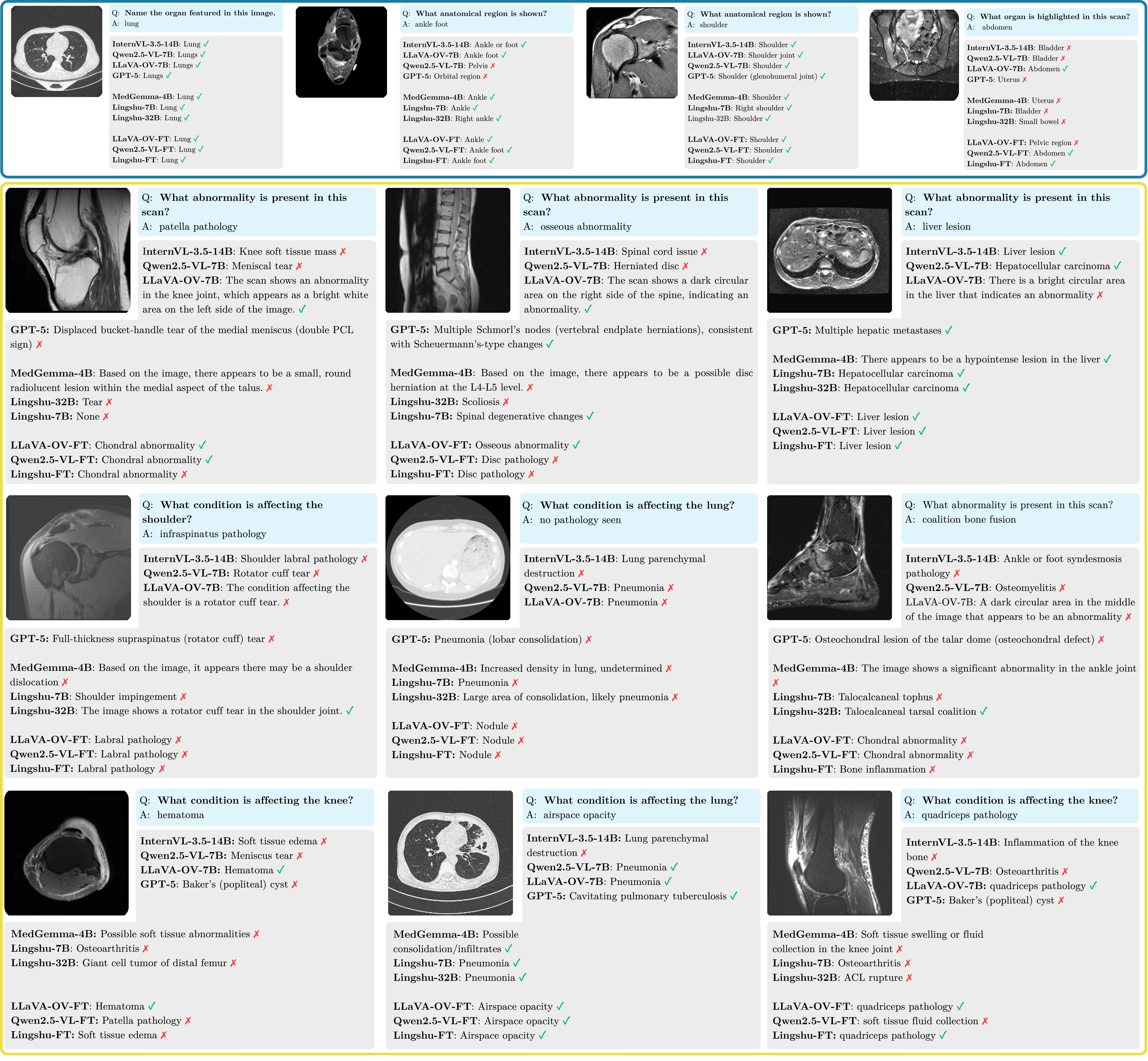}
          \vspace{-0.5cm}
    \caption{Qualitative examples of model predictions for anatomy recognition (\textit{top}) and pathology identification (\textit{bottom}). Each row shows one case with ground-truth; predictions are marked as correct (\cmark) or incorrect (\xmark) according to LLM-Judge.}
    \label{fig:anat_path_qualitative_examples}
        \vspace{-0.5cm}
\end{figure}
}

\subsection{Qualitative Analysis}

To better understand model behavior on open-ended questions, Figure~\ref{fig:anat_path_qualitative_examples} presents some qualitative examples comparing anatomy recognition and pathology identification. Anatomy recognition is generally robust across models, especially for large, well-defined structures (e.g., shoulder, lung, ankle/foot) where shape and contextual cues are strong. For broader regions such as the abdomen, zero-shot models sometimes predict a salient sub-organ (e.g., bladder/uterus) rather than the full region.

In contrast, pathology identification remains the primary bottleneck. Models frequently misidentify or overlook subtle, localized, or rare abnormalities. For example, small or fine-grained lesions such as  \textit{patella pathology}, \textit{coalition bone fusion}, or \textit{quadriceps pathology} are missed or confuse with surrounding structures by many zero-shot models (e.g., GPT-5, MedGemma-4B, Lingshu-32B). Similarly, \textit{osseous abnormalities} in the spine or \textit{infraspinatus/rotator cuff injuries} were often misclassified or described ambiguously, highlighting texture ambiguity and small-object difficulty. Even visually salient findings such as \textit{liver lesions} or \textit{airspace opacities} can be inconsistently labeled across models, suggesting sensitivity to appearance variation. Fine-tuning improves recognition for some categories (e.g., \textit{chondral abnormalities}, \textit{liver lesions}), but rare or subtle pathologies like \textit{Lisfranc ligament injury}, and \textit{ACL tears}, or\textit{coalition bone fusion} remain challenging.

%==================================================================

\section{Discussion}
RadImageNet-VQA provides large-scale CT and MRI supervision for radiologic VQA, spanning abnormality detection, anatomy recognition, and fine-grained pathology identification across multiple question formats. Benchmark results show that current VLMs perform relatively well on anatomy and abnormality recognition, while fine-grained pathology identification remains the main failure mode even after fine-tuning. Together with the text-only ablations, these results suggest that progress will require better visual grounding for subtle findings rather than exploiting linguistic priors. Beyond benchmarking, the dataset serves as a diagnostic tool for model behavior and a scalable fine-tuning resource for VLMs. Additionally, the inclusion of both CT and MRI also enables future cross-modality studies.

\paragraph{Limitations.} The dataset inherits several constraints from the underlying RadImageNet taxonomy, most notably a single-label pathology per image~\cite{mei2022radimagenet}. This setting does not fully reflect the complexity of some real-world clinical cases which can involve multiple co-existing findings. Hence, promising extensions include adding samples with multi-finding annotations and richer clinical context. Potential biases such as pathology prevalence and imaging protocols may also disadvantage underrepresented conditions. Beyond these inherited aspects, our formulation has additional limitations: (1) questions are programmatically generated, which increases control and scalability but may under-represent real clinical language and reasoning; (2) the dataset is 2D, so models do not have access to volumetric context; (3) open-ended evaluation relies on an automatic judge, which we validate but may still introduce residual uncertainty for borderline cases.

%==================================================================
\section{Conclusion}

In this work, we introduce RadImageNet-VQA, a large-scale CT/MRI dataset for radiologic VQA built through a rigorous QA-generation pipeline. It includes both a training set and a curated benchmark, covering multiple tasks and anatomy regions. Our experiments show that fine-grained pathology identification remains challenging for current models. Fine-tuning current VLMs on RadImageNet-VQA yields substantial performance gains, while greatly reducing text-based shortcuts, which demonstrate its potential to be a valuable resource for developing and assessing stronger medical VLMs for radiology.

\section*{Acknowledgment}
We thank the RadImageNet team for granting access to the RadImageNet dataset and for making this resource available for research purposes (\url{https://www.radimagenet.com/}). 
We thank the Jean Zay and Leonardo HPC centers for providing the GPU computing resources used in this work.

\bibliography{midl26_280}

\clearpage

%==================================================================
%=========                   APPENDIX                   ===========
%==================================================================

\appendix

%==================================================================
\section{Dataset details}
\label{appx_sec:dataset-details}

%------------------------------------------------------------------
\subsection{RadImageNet-VQA anatomy regions and pathology taxonomy}
\label{appx_subsec:detailed_anatomy_pathologies}

The full taxonomy of CT and MRI anatomical regions paired with fine-grained pathology labels used for caption and VQA generation is reported in Table~\ref{appx_tab:anatomy_pathologies}.

\begin{table}[h]
\centering
\label{tab:anatomy_pathologies}
\footnotesize
\begin{tabular}{>{\raggedright\arraybackslash}p{0.18\textwidth} >{\raggedright\arraybackslash}p{0.77\textwidth}}
\toprule
\textbf{Anatomy} & \textbf{Pathologies} \\
\midrule
abdomen & 
abnormal entire organ, adrenal pathology, arterial pathology, ascites, biliary dilation, bladder pathology, bowel abnormality, bowel inflammation, bowel mass, degenerative changes, dilated urinary tract, enlarged organ, fat-containing tumor, gallbladder pathology, gallstone, intraperitoneal mass, liver disease, liver lesion, marrow abnormality, osseous neoplasm, ovarian pathology, pancreatic lesion, post-operative state, prostate lesion, renal lesion, soft tissue collection, soft tissue mass, splenic lesion, urolithiasis, uterine pathology \\
\addlinespace

ankle foot & 
achilles pathology, anterior talofibular ligament pathology, bone inflammation, calcaneofibular ligament pathology, chondral abnormality, coalition (bone fusion), deltoid pathology, extensor pathology, fat-containing tumor, flexor pathology, hematoma, intra-articular abnormality, Lisfranc joint injury, osseous disruption, osseous neoplasm, peroneal tendon pathology, plantar fascia pathology, plantar plate tear, post-operative state, soft tissue edema, soft tissue fluid, soft tissue mass, spring ligament injury, syndesmosis pathology \\
\addlinespace

brain & 
acute infarct, arteriovenous anomaly, chronic infarct, edema, extra-axial lesion, focal FLAIR hyperintensity, intra-articular abnormality, pituitary lesion, white matter changes \\
\addlinespace

hip & 
abductor pathology, capsular pathology, chondral pathology, hamstring pathology, hematoma, labral pathology, marrow inflammation, osseous disruption, osseous lesion, post-operative state, soft tissue edema, soft tissue fluid, soft tissue mass \\
\addlinespace

knee & 
anterior cruciate ligament pathology, bone inflammation, chondral abnormality, fibular collateral ligament pathology, fracture, hematoma, intra-articular abnormality, medial collateral ligament pathology, meniscal abnormality, muscle strain, patella pathology, post-operative ACL reconstruction, posterior cruciate ligament pathology, quadriceps pathology, soft tissue edema, soft tissue fluid collection, soft tissue mass \\
\addlinespace

lung & 
airspace opacity, bronchiectasis, interstitial lung disease, nodule, parenchyma destruction \\
\addlinespace

shoulder & 
acromioclavicular joint osteoarthritis, biceps pathology, calcific tendinosis, glenohumeral joint osteoarthritis, infraspinatus pathology, labral pathology, marrow inflammation, osseous lesion, post-operative state, soft tissue edema, soft tissue fluid, subscapularis pathology, supraspinatus pathology \\
\addlinespace

spine & 
cystic lesions, disc pathology, dural or epidural abnormality, facet arthropathy, foraminal pathology, osseous abnormality, scoliosis, spinal cord pathology \\
\bottomrule
\end{tabular}
\caption{Anatomical regions and corresponding pathologies in RadImageNet-VQA}
\label{appx_tab:anatomy_pathologies}
\end{table}

\subsection{RadImageNet-VQA anatomy regions and pathology distribution}

\begin{table}[H]
\centering
\caption{\textbf{Anatomy region distribution} in train corpus and benchmark, sorted by decreasing training-set frequency. Percentages are computed over images. $\Delta$pp are percentages differences in points of benchmark compared to train.}
\label{tab:anatomy_dist}
\begin{tabular}{lrrr}
\toprule
\textbf{Anatomy} & \textbf{Train count (\%)} & \textbf{Benchmark count (\%)} & \textbf{$\Delta$}pp \\
\midrule
Abdomen     & 189,104 (25.21\%) & 313 (31.30\%) & +6.09 \\
Ankle Foot  & 146,009 (19.47\%) & 104 (10.40\%) & -9.07 \\
Knee        & 145,493 (19.40\%) & 87 ~(8.70\%)  & -10.70 \\
Lung        &  89,709 (11.96\%) & 193 (19.30\%) & +7.34 \\
Spine       &  57,038 ~(7.60\%)  & 75 ~(7.50\%)  & -0.10 \\
Shoulder    &  42,669 ~(5.69\%)  & 77 ~(7.70\%)  & +2.01 \\
Hip         &  42,480 ~(5.66\%)  & 74 ~(7.40\%) & +1.74 \\
Brain       &  37,507 ~(5.00\%)  & 77 ~(7.70\%)  & +2.70 \\
\bottomrule
\end{tabular}
\end{table}

\begingroup
%\footnotesize
\setlength{\tabcolsep}{4pt}
\renewcommand{\arraystretch}{0.95}

\begin{longtable}{lrrr}
\caption{\textbf{Pathology distribution.} in train corpus and benchmark, , sorted by decreasing training-set frequency. Percentages are computed over images. $\Delta$pp are percentages differences in points of benchmark compared to train..}
\label{tab:pathology_distribution_full} \\
\toprule
\textbf{Pathology} & \textbf{Train count (\%)} & \textbf{Benchmark count (\%)} & \textbf{$\Delta$pp} \\
\midrule
\endfirsthead

\toprule
\textbf{Pathology} & \textbf{Train count (\%)} & \textbf{Benchmark count (\%)} & \textbf{$\Delta$pp} \\
\midrule
\endhead

\midrule
\multicolumn{4}{r}{Continued on next page} \\
\midrule
\endfoot

\bottomrule
\endlastfoot

Normal & 196,507 (26.20\%) & 281 (28.10\%) & +1.90 \\
Chondral Abnormality & 76,222 (10.16\%) & 33 (3.30\%) & -6.86 \\
Meniscal Abnormality & 36,414 (4.86\%) & 15 (1.50\%) & -3.36 \\
Soft Tissue Fluid & 33,461 (4.46\%) & 25 (2.50\%) & -1.96 \\
Labral Pathology & 32,744 (4.37\%) & 38 (3.80\%) & -0.57 \\
Bone Inflammation & 32,508 (4.33\%) & 13 (1.30\%) & -3.03 \\
Airspace Opacity & 30,861 (4.11\%) & 50 (5.00\%) & +0.89 \\
Disc Pathology & 26,415 (3.52\%) & 28 (2.80\%) & -0.72 \\
Nodule & 21,047 (2.81\%) & 28 (2.80\%) & -0.01 \\
Soft Tissue Fluid Collection & 20,781 (2.77\%) & 14 (1.40\%) & -1.37 \\
Soft Tissue Edema & 14,178 (1.89\%) & 12 (1.20\%) & -0.69 \\
Liver Lesion & 11,714 (1.56\%) & 20 (2.00\%) & +0.44 \\
Renal Lesion & 11,275 (1.50\%) & 17 (1.70\%) & +0.20 \\
Osseous Disruption & 10,655 (1.42\%) & 7 (0.70\%) & -0.72 \\
ATFL Pathology & 9,698 (1.29\%) & 6 (0.60\%) & -0.69 \\
Foraminal Pathology & 9,674 (1.29\%) & 9 (0.90\%) & -0.39 \\
Post-operative Changes & 9,009 (1.20\%) & 11 (1.10\%) & -0.10 \\
ACL Pathology & 8,191 (1.09\%) & 4 (0.40\%) & -0.69 \\
White Matter Changes & 8,166 (1.09\%) & 18 (1.80\%) & +0.71 \\
Bowel Abnormality & 7,966 (1.06\%) & 14 (1.40\%) & +0.34 \\
Interstitial Lung Disease & 7,872 (1.05\%) & 8 (0.80\%) & -0.25 \\
Soft Tissue Mass & 7,763 (1.04\%) & 6 (0.60\%) & -0.44 \\
Achilles Pathology & 7,413 (0.99\%) & 4 (0.40\%) & -0.59 \\
Dural/Epidural Abnormality & 6,668 (0.89\%) & 10 (1.00\%) & +0.11 \\
Chondral Pathology & 6,598 (0.88\%) & 15 (1.50\%) & +0.62 \\
MCL Pathology & 6,427 (0.86\%) & 4 (0.40\%) & -0.46 \\
Peroneal Pathology & 6,017 (0.80\%) & 4 (0.40\%) & -0.40 \\
Bowel Inflammation & 5,910 (0.79\%) & 17 (1.70\%) & +0.91 \\
Intracranial Pathology & 5,338 (0.71\%) & 5 (0.50\%) & -0.21 \\
Uterine Pathology & 5,196 (0.69\%) & 12 (1.20\%) & +0.51 \\
Plantar Fascia Pathology & 5,183 (0.69\%) & 6 (0.60\%) & -0.09 \\
Supraspinatus Pathology & 4,197 (0.56\%) & 5 (0.50\%) & -0.06 \\
Marrow Inflammation & 3,853 (0.51\%) & 7 (0.70\%) & +0.19 \\
CFL Pathology & 3,395 (0.45\%) & 4 (0.40\%) & -0.05 \\
Osseous Neoplasm & 3,324 (0.44\%) & 6 (0.60\%) & +0.16 \\
Flexor Pathology & 2,915 (0.39\%) & 4 (0.40\%) & +0.01 \\
Pancreatic Lesion & 2,646 (0.35\%) & 7 (0.70\%) & +0.35 \\
Ovarian Pathology & 2,633 (0.35\%) & 5 (0.50\%) & +0.15 \\
Glenohumeral Joint OA & 2,378 (0.32\%) & 5 (0.50\%) & +0.18 \\
Deltoid Pathology & 2,209 (0.29\%) & 4 (0.40\%) & +0.11 \\
Osseous Lesion & 2,174 (0.29\%) & 6 (0.60\%) & +0.31 \\
Acromioclavicular Joint OA & 2,158 (0.29\%) & 3 (0.30\%) & +0.01 \\
Adrenal Pathology & 2,101 (0.28\%) & 5 (0.50\%) & +0.22 \\
Gallstone & 1,923 (0.26\%) & 7 (0.70\%) & +0.44 \\
Osseous Abnormality & 1,911 (0.25\%) & 3 (0.30\%) & +0.05 \\
Ascites & 1,853 (0.25\%) & 8 (0.80\%) & +0.55 \\
Chronic Infarct & 1,842 (0.25\%) & 7 (0.70\%) & +0.45 \\
Scoliosis & 1,655 (0.22\%) & 4 (0.40\%) & +0.18 \\
Bladder Pathology & 1,599 (0.21\%) & 5 (0.50\%) & +0.29 \\
Parenchymal Destruction & 1,519 (0.20\%) & 5 (0.50\%) & +0.30 \\
Bronchiectasis & 1,408 (0.19\%) & 4 (0.40\%) & +0.21 \\
Urolithiasis & 1,378 (0.18\%) & 8 (0.80\%) & +0.62 \\
Biceps Pathology & 1,323 (0.18\%) & 5 (0.50\%) & +0.32 \\
Intraperitoneal Mass & 1,303 (0.17\%) & 3 (0.30\%) & +0.13 \\
Fracture & 1,275 (0.17\%) & 3 (0.30\%) & +0.13 \\
Arterial Pathology & 1,154 (0.15\%) & 6 (0.60\%) & +0.45 \\
Patella Pathology & 1,147 (0.15\%) & 3 (0.30\%) & +0.15 \\
Cystic Lesions & 1,104 (0.15\%) & 3 (0.30\%) & +0.15 \\
Extrastructural Pathology & 1,020 (0.14\%) & 7 (0.70\%) & +0.56 \\
Dilated Urinary Tract & 955 (0.13\%) & 8 (0.80\%) & +0.67 \\
Liver Disease & 868 (0.12\%) & 3 (0.30\%) & +0.18 \\
Prostate Lesion & 835 (0.11\%) & 3 (0.30\%) & +0.19 \\
Biliary Dilation & 833 (0.11\%) & 3 (0.30\%) & +0.19 \\
Abductor Pathology & 724 (0.10\%) & 3 (0.30\%) & +0.20 \\
Splenic Lesion & 718 (0.10\%) & 3 (0.30\%) & +0.20 \\
Hematoma & 642 (0.09\%) & 4 (0.40\%) & +0.31 \\
Marrow Abnormality & 600 (0.08\%) & 3 (0.30\%) & +0.22 \\
Focal FLAIR Hyperintensity & 592 (0.08\%) & 6 (0.60\%) & +0.52 \\
PCL Pathology & 585 (0.08\%) & 4 (0.40\%) & +0.32 \\
Spinal Cord Pathology & 580 (0.08\%) & 3 (0.30\%) & +0.22 \\
Fat-containing Tumor & 540 (0.07\%) & 3 (0.30\%) & +0.23 \\
Plantar Plate Tear & 536 (0.07\%) & 3 (0.30\%) & +0.23 \\
Extensor Pathology & 536 (0.07\%) & 3 (0.30\%) & +0.23 \\
Gallbladder Pathology & 517 (0.07\%) & 3 (0.30\%) & +0.23 \\
Bowel Mass & 460 (0.06\%) & 3 (0.30\%) & +0.24 \\
Muscle Strain & 424 (0.06\%) & 3 (0.30\%) & +0.24 \\
Acute Infarct & 423 (0.06\%) & 3 (0.30\%) & +0.24 \\
Quadriceps Pathology & 405 (0.05\%) & 3 (0.30\%) & +0.25 \\
FCL Pathology & 367 (0.05\%) & 3 (0.30\%) & +0.25 \\
Syndesmosis Pathology & 318 (0.04\%) & 3 (0.30\%) & +0.26 \\
Facet Arthropathy & 302 (0.04\%) & 3 (0.30\%) & +0.26 \\
Soft Tissue Collection & 238 (0.03\%) & 3 (0.30\%) & +0.27 \\
Arteriovenous Anomaly & 217 (0.03\%) & 3 (0.30\%) & +0.27 \\
Global Organ Abnormality & 211 (0.03\%) & 3 (0.30\%) & +0.27 \\
Degenerative Changes & 197 (0.03\%) & 3 (0.30\%) & +0.27 \\
Hamstring Pathology & 192 (0.03\%) & 3 (0.30\%) & +0.27 \\
Organomegaly & 170 (0.02\%) & 3 (0.30\%) & +0.28 \\
Subscapularis Pathology & 115 (0.02\%) & 3 (0.30\%) & +0.28 \\
Capsular Pathology & 99 (0.01\%) & 3 (0.30\%) & +0.29 \\
Edema & 98 (0.01\%) & 3 (0.30\%) & +0.29 \\
Spring Ligament Injury & 83 (0.01\%) & 3 (0.30\%) & +0.29 \\
Infraspinatus Pathology & 76 (0.01\%) & 3 (0.30\%) & +0.29 \\
Post-operative ACL & 71 (0.01\%) & 3 (0.30\%) & +0.29 \\
Pituitary Lesion & 64 (0.01\%) & 3 (0.30\%) & +0.29 \\
Calcific Tendinosis & 63 (0.01\%) & 3 (0.30\%) & +0.29 \\
Tarsal Coalition & 50 (0.01\%) & 3 (0.30\%) & +0.29 \\
Lisfranc Pathology & 37 (0.00\%) & 3 (0.30\%) & +0.30 \\

\end{longtable}
\endgroup

\begin{figure}[H]

    \centering
    \includegraphics[width=1
\linewidth]{graphs/train-distrib.png}
    \vspace{-0.5cm}
    \caption{\textbf{Composition of the RadImageNet-VQA training set}, containing 750k CT/MRI images : 
(a) most frequent pathology labels, 
(b) less frequent pathology labels, 
(c) anatomy distribution.
Percentages are computed over images.
}
    \vspace{-0.6cm}
    \label{fig:train_distrib}
\end{figure}

\newpage
%------------------------------------------------------------------
\subsection{Template set for VQA and captioning generation}
\label{appx_subsec:question_templates}

\begin{figure}[H]
    \vspace{-0.7cm}
    \centering
    \includegraphics[width=0.8\linewidth]{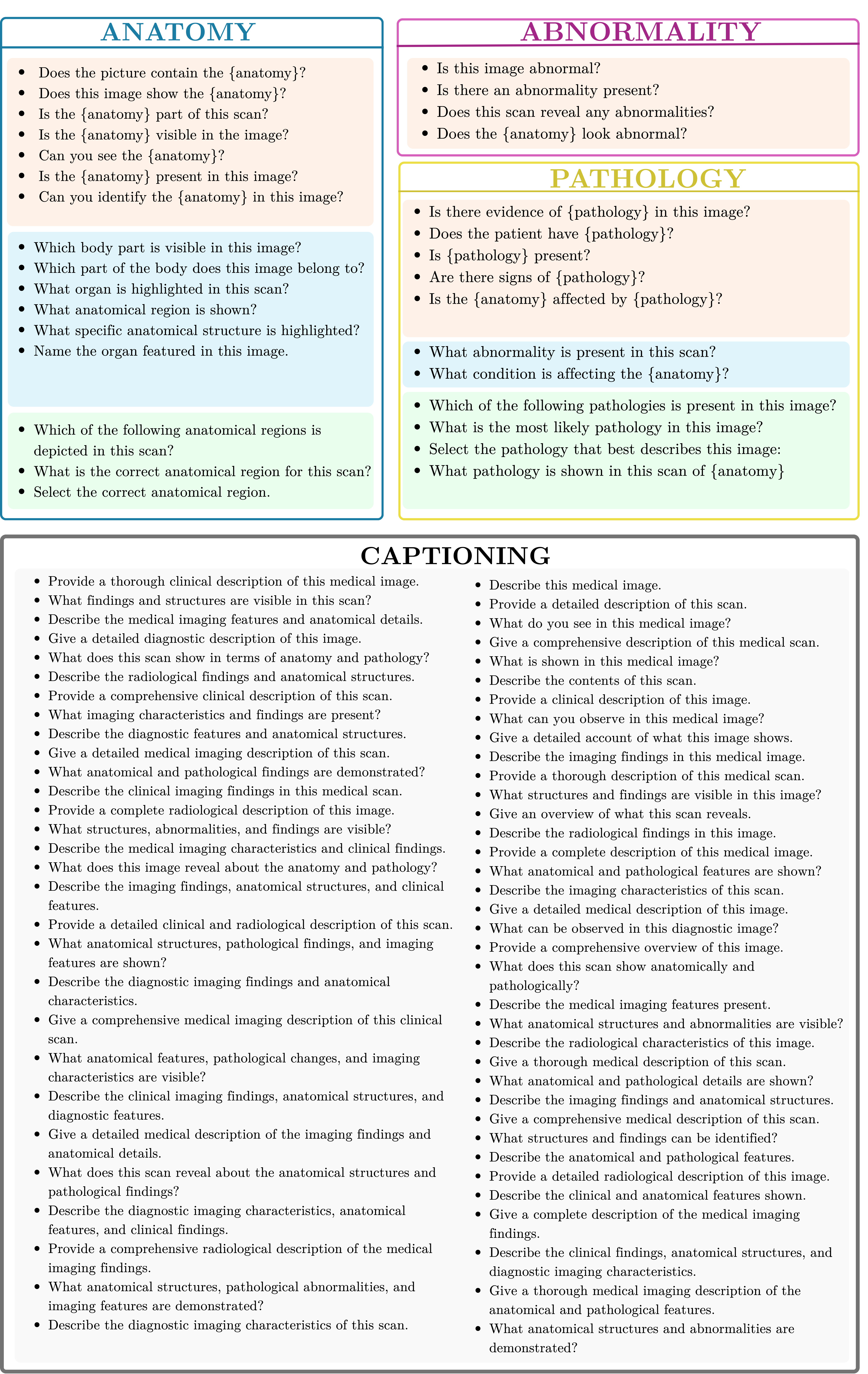}
    \vspace{-0.4cm}
    \caption{Templates of questions for VQA samples and radiologic captions.}
    \label{appx_fig:question-templates}
\end{figure}

Figure~\ref{appx_fig:question-templates} illustrates the template families used to generate radiologic captions and VQA samples for all tasks and question types.

\section{LLM-as-a-judge evaluation}
\label{appx_sec:llm_judge}

\subsection{Protocol of evaluation}
\label{appx_subsec:llm-judge-prompt}

We adopt the MedEvalKit framework\footnote{\url{https://github.com/alibaba-damo-academy/MedEvalKit}}
 released with Lingshu-7B~\cite{xu2025lingshu} and use Mistral-Large~2.1~\cite{mistral2024large2} as the judge model.

\begin{figure}[H]
\centering
\begin{mdframed}[backgroundcolor=grey!10]
\scriptsize
\ttfamily
\sloppy
Your task is to determine whether the user’s answer is correct based on the provided questions and standard answers (for example, if the user expresses a similar meaning to the standard answer, or another interpretation of the standard answer, it is considered correct.)

The question is: \{question\}

The standard answer: \{answer\} 
The user’s answer: \{response\}

Please strictly follow the following format for output (0 represents correct, 1 represents incorrect):
<think> your concise think step </think> \quad
<judge>0/1</judge>

For example:

<think>
The standard answer is right, and the user’s answer is right frontal lobe; they express the
same meaning, so it is correct.
</think>
<judge>0</judge>

\label{fig:llm-judge-prompt}
\end{mdframed}
\vspace{-0.4cm}
\caption{Judgment prompt used in the LLM-as-a-judge module.}
\end{figure}

\subsection{Validation of LLM-judge}
\label{appx_subsec:llm_judge_validation}

Table~\ref{tab:llm_judge_agreement} presents the validation results where two independent reviewers evaluated 380 stratified samples. Reviewers could mark cases as \emph{correct}, \emph{incorrect}, \emph{don't know}, or \emph{response more precise than ground truth}. For the reported metrics, cases marked as \emph{ response more precise than ground truth} were treated as correct, while \emph{ don't know} cases ($\sim$6\% of total) were conservatively treated as incorrect.

\begin{table}[h]
\centering
\begin{tabular}{lccc}
\hline
\multirow{2}{*}{\textbf{Reviewer}} & \multicolumn{2}{c}{\textbf{Agreement (\%)}} & \multirow{2}{*}{\textbf{Cohen's $\kappa$}} \\
\cline{2-3}
& Anatomy & Pathology & \\
\hline
Reviewer 1 & 94.20\% & 81.05\% & 0.7273 \\
Reviewer 2 & 93.98\% & 77.27\% & 0.7135 \\
\hline
\end{tabular}

\caption{Validation of LLM judge against human reviewers on 380 random samples from various models.}

\label{tab:llm_judge_agreement}
\end{table}

%==================================================================

%==================================================================
\section{Additional Zero-Shot Analysis}
\label{appx_sec:add-exp}

%------------------------------------------------------------------
\subsection{Effect of ``no pathology seen'' distractors in pathology MC questions}
\label{appx_subsec:add-distractors}

To assess how the construction of multiple-choice distractors influences model behavior on RadImageNet-VQA, we vary the proportion of pathology questions whose answer set includes the option ``no pathology seen''. We evaluate four representative VLMs—two medically oriented models (Lingshu-7B, MedGemma-4B-it) and two general-purpose models (InternVL3.5-8B, Qwen2.5-VL-7B)—under three configurations in which this option appears in 0\%, 30\%, or 100\% of pathology multiple-choice questions.

\begin{wraptable}{r}{0.5\textwidth}
\centering
\vspace{-0.3cm}
\renewcommand{\arraystretch}{1.2}
\setlength{\tabcolsep}{3pt}
\scriptsize
\begin{tabular}{l|ccc}
\toprule
\textbf{Model} & \textbf{0\%} & \textbf{30\%} & \textbf{100\%} \\
\midrule
\rowcolor{gray!10}
\multicolumn{4}{l}{\textbf{General-purpose models}} \\
InternVL3.5\_8B & 45.0 & 44.7 & 46.5 \\
Qwen2.5-VL-7B & 40.5 & 40.3 & 42.6 \\
\midrule
\rowcolor{gray!10}
\multicolumn{4}{l}{\textbf{Medical specialized models}} \\
Lingshu-7B & 59.2 & 54.9 & 49.1 \\
medgemma-4b-it & 60.9 & 57.6 & 50.2 \\
\bottomrule
\end{tabular}
\caption{Accuracy on pathology MC questions when varying the frequency of the ``no pathology seen'' option.}
\label{appx_tab:unified_accuracy}
\vspace{-0.4cm}
\end{wraptable}

As reported in Table~\ref{appx_tab:unified_accuracy}, MedGemma and Lingshu show a progressive decrease in accuracy as the “no pathology’’ choice becomes more frequent, suggesting that they tend to assume the presence of pathology even when presented with an explicit normal alternative. In contrast, InternVL and Qwen remain comparatively stable or slightly improve, indicating that general models are more inclined to select this choice when uncertain or to rely on deduction from the visible options. Overall, these results show that distractor design can substantially alter model responses and that explicitly modelling the possibility of normal findings remains important for robust radiologic VQA.

\begin{figure}[t]
    \centering
    \includegraphics[width=1\linewidth]{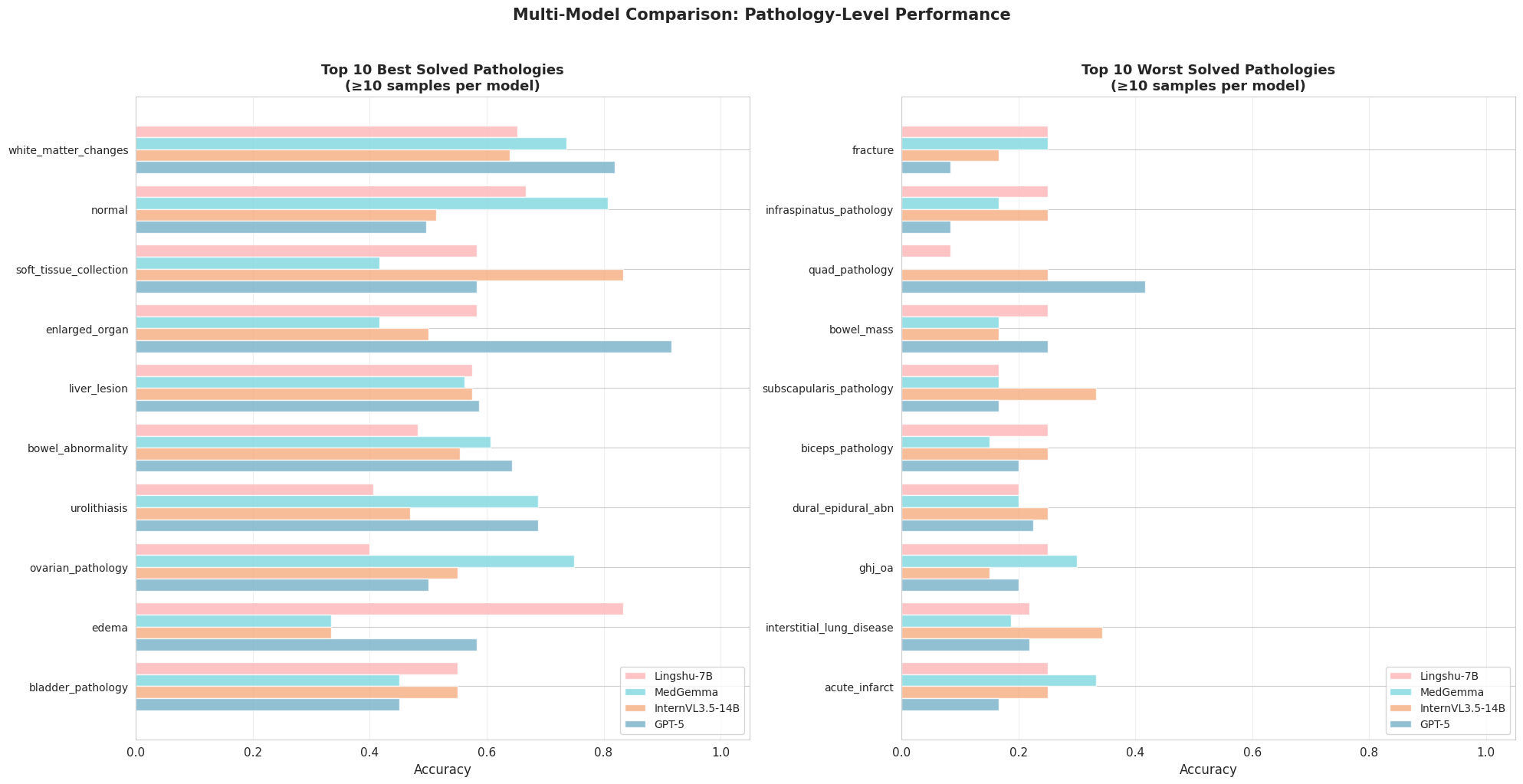}
    \caption{Per-pathology multiple-choice accuracy across four VLMs on RadImageNet-VQA. \textit{Left}: 10 best-solved pathologies; \textit{right}: 10 worst-solved}
    \label{appx_fig:path-accuracies}
\end{figure}

%------------------------------------------------------------------
\subsection{Per-pathology performance analysis}
\label{appx_subsec:zero-shot}

We observe in Table~\ref{tab:zero-shot} that pathology identification is the most challenging task in the zero-shot setting; to further understand which specific conditions drive this difficulty, Figure~\ref{appx_fig:path-accuracies} reports pathology-level performance by listing the ten best- and worst-solved categories. We find that common and visually salient abnormalities (e.g., organ enlargement, soft-tissue collections) are generally solved reliably, whereas fine-grained or anatomically localized findings—particularly musculoskeletal pathologies—remain difficult for all models. Although medically oriented models perform competitively on frequent categories, they do not consistently outperform general-purpose models on the harder ones, and InternVL often achieves the best results overall, suggesting that detailed disease characterization rather than abnormality detection constitutes the main bottleneck.

{

\subsection{Robustness to question template variations}
\label{appx:template_sensitivity}

To assess whether models exploit linguistic shortcuts or are overly sensitive to question phrasing, we analyze model performance across different natural-language templates within the same task, category, and question type.
Each template corresponds to a distinct formulation of an equivalent semantic query (e.g., abnormality detection or pathology presence).

Table~\ref{tab:results_per_template} reports accuracy for each model across all templates.
Templates are well distributed within each task category, and no single phrasing dominates the benchmark.
We note that different templates are instantiated on different image subsets; therefore, small variations in accuracy may partially reflect sample variability rather than purely linguistic effects.

Overall, model performance remains relatively stable across paraphrased templates within the same task, suggesting that the benchmark does not contain strong template-specific shortcuts. This is particularly evident in anatomical region and structure identification tasks, where accuracy varies only marginally across formulations. However, higher sensitivity is observed for certain models in linguistically challenging settings, notably open-ended pathology identification and closed-form pathology detection on negative cases. For example, in open-ended pathology identification, MedGemma varies from 26.28\% to 34.70\% across paraphrased templates, while InternVL3.5 drops from 14.99\% to 8.58\%. In negative pathology detection, InternVL3.5 ranges from 48.39\% to 7.71\% and GPT-5 from 73.66\% to 8.44\% depending on phrasing, whereas MedGemma and Lingshu remain comparatively stable (typically $>$75\%). These variations indicate model-specific sensitivity to wording rather than systematic bias in the benchmark templates.

Taken together, these results support the robustness of our question templating strategy while revealing meaningful differences in how models handle linguistic variation.

\begin{table}[t]
\centering
\caption{\textbf{Per-template accuracy (\%)} across models, grouped by question type (Closed+/Closed-/Open-ended/MCQ). Count denotes the number of benchmark questions instantiating each template.}
\label{tab:results_per_template}
\setlength{\tabcolsep}{2pt}
\renewcommand{\arraystretch}{1.05}
\resizebox{\linewidth}{!}{%
\begin{tabular}{lccccccc}
\toprule
\textbf{Question Template} & \textbf{Count} & \textbf{InternVL} & \textbf{Qwen-VL} & \textbf{GPT-5} & \textbf{MedGemma} & \textbf{Lingshu} \\
\midrule
\rowcolor{gray!10}
\multicolumn{7}{l}{\textit{Abnormality Detection - Closed }} \\
Does the \{anatomy\} look abnormal? & 266 & 77.07 & 72.18 & 25.94 & 60.90 & 50.38 \\
Does this scan reveal any abnormalities? & 220 & 76.36 & 58.64 & 18.18 & 56.36 & 39.55 \\
Is there an abnormality present? & 256 & 73.05 & 72.27 & 44.53 & 55.86 & 51.95 \\
Is this image abnormal? & 258 & 71.32 & 73.26 & 20.16 & 48.45 & 48.45 \\
\midrule

\rowcolor{gray!10}
\multicolumn{7}{l}{\textit{Pathology Identification - Open-ended}} \\
What abnormality is present? & 487 & 14.99 & 10.47 & 18.69 & 26.28 & 15.61 \\
What condition is affecting the anatomy? & 513 & 8.58 & 9.16 & 13.06 & 34.70 & 15.79 \\
\midrule

\rowcolor{gray!10}
\multicolumn{7}{l}{\textit{Pathology Detection - Closed  (Positive Cases)}} \\
Are there signs of \{pathology\}? & 196 & 84.69 & 65.82 & 51.02 & 48.98 & 54.08 \\
Does the patient have \{pathology\}? & 195 & 87.18 & 69.23 & 57.44 & 58.97 & 65.64 \\
Is the \{anatomy\} affected by \{pathology\}? & 190 & 88.42 & 67.89 & 54.21 & 53.16 & 58.42 \\
Is there evidence of \{pathology\} in this image? & 218 & 86.24 & 73.39 & 53.21 & 52.29 & 51.38 \\
Is \{pathology\} present? & 201 & 87.06 & 69.15 & 50.75 & 57.71 & 56.22 \\
\midrule
\rowcolor{gray!10}
\multicolumn{7}{l}{\textit{Pathology Detection - Closed  (Negative Cases)}} \\
Are there signs of \{pathology\}? & 186 & 48.39 & 53.23 & 73.66 & 81.72 & 84.95 \\
Does the patient have \{pathology\}? & 206 & 22.33 & 38.83 & 59.71 & 67.96 & 66.02 \\
Is the \{anatomy\} affected by \{pathology\}? & 200 & 18.50 & 52.00 & 61.00 & 77.00 & 80.00 \\
Is there evidence of \{pathology\} in this image? & 218 & 7.71 & 42.20 & 8.44 & 84.86 & 83.03 \\
Is \{pathology\} present? & 190 & 31.58 & 52.11 & 68.42 & 75.26 & 80.53 \\
\midrule
\rowcolor{gray!10}
\multicolumn{7}{l}{\textit{Pathology Classification - Multiple Choice}} \\
Select the pathology that best describes this image & 219 & 47.03 & 27.85 & 42.92 & 31.05 & 25.11 \\
What is the most likely pathology in this image? & 260 & 46.15 & 31.92 & 40.77 & 36.54 & 29.62 \\
What pathology is shown in this scan of \{anatomy\}? & 250 & 48.80 & 32.40 & 39.60 & 36.80 & 28.80 \\
Which of the following pathologies is present in this image? & 271 & 46.49 & 28.04 & 41.70 & 41.70 & 33.95 \\
\midrule
\rowcolor{gray!10}
\multicolumn{7}{l}{\textit{Anatomical Region Identification - Open-ended}} \\
Name the organ featured in this image & 177 & 49.72 & 25.99 & 36.16 & 54.24 & 55.93 \\
What anatomical region is shown? & 162 & 65.43 & 46.30 & 58.64 & 69.75 & 54.94 \\
What organ is highlighted in this scan? & 183 & 53.01 & 28.42 & 34.97 & 66.12 & 45.36 \\
What specific anatomical structure is highlighted? & 172 & 39.53 & 22.09 & 8.14 & 51.74 & 30.81 \\
Which body part is visible in this image? & 156 & 68.59 & 58.97 & 64.10 & 75.00 & 60.26 \\
Which part of the body does this image belong to? & 150 & 66.67 & 48.00 & 70.67 & 62.00 & 52.00 \\
\midrule
\rowcolor{gray!10}
\multicolumn{7}{l}{\textit{Anatomical Structure Presence Detection - Closed}} \\
Can you identify the \{anatomy\} in this image? & 146 & 82.18 & 80.77 & 30.94 & 48.82 & 86.63 \\
Can you see the \{anatomy\}? & 136 & 88.49 & 85.75 & 86.96 & 85.84 & 90.33 \\
Does the picture contain the \{anatomy\}? & 139 & 88.96 & 81.73 & 86.43 & 84.81 & 88.79 \\
Does this image show the \{anatomy\}? & 145 & 88.38 & 80.62 & 82.92 & 87.30 & 86.83 \\
Is the \{anatomy\} part of this scan? & 139 & 84.69 & 79.94 & 80.28 & 80.83 & 85.89 \\
Is the \{anatomy\} present in this image? & 145 & 87.46 & 84.84 & 85.84 & 84.83 & 89.70 \\
Is the \{anatomy\} visible in the image? & 152 & 83.78 & 80.13 & 87.18 & 83.90 & 86.94 \\
\midrule
\rowcolor{gray!10}
\multicolumn{7}{l}{\textit{Anatomical Region Selection - Multiple Choice}} \\
Select the correct anatomical region & 342 & 88.30 & 79.82 & 88.60 & 83.33 & 87.13 \\
What is the correct anatomical region for this scan? & 321 & 88.79 & 79.44 & 88.47 & 82.87 & 89.10 \\
Which of the following anatomical regions is depicted in this scan? & 337 & 92.58 & 82.20 & 90.80 & 88.13 & 90.50 \\
\bottomrule
\end{tabular}
}
\end{table}

\subsection{Out-of-template evaluation of fine-tuned models}

While RadImageNet-VQA provides carefully designed question templates to reduce superficial language biases, these templated formats may inadvertently encourage models to rely on repeated linguistic patterns rather than true visual understanding. To assess the robustness of our fine-tuned models beyond these constraints, we conducted an out-of-template evaluation, in which questions were rephrased or presented in styles not seen during training. This setup simulates realistic clinical queries, including open-ended descriptions and higher-level reasoning, that go beyond the template structures seen in training (e.g., asking “What do you see in the heart?” instead of “Which body part is visible in this image?”).

For this experiment, we tested three representative CT/MRI images: a shoulder scan with soft-tissue fluid, a spine scan with foraminal pathology, and an abdominal scan with uterine pathology. Questions were designed to probe both anatomical identification and pathology interpretation, including general prompts, localization questions, and complication reasoning. The qualitative results are summarized in Figure~\ref{fig:out-of-template}.

\begin{figure}[H]
\centering
\includegraphics[width=1\linewidth]{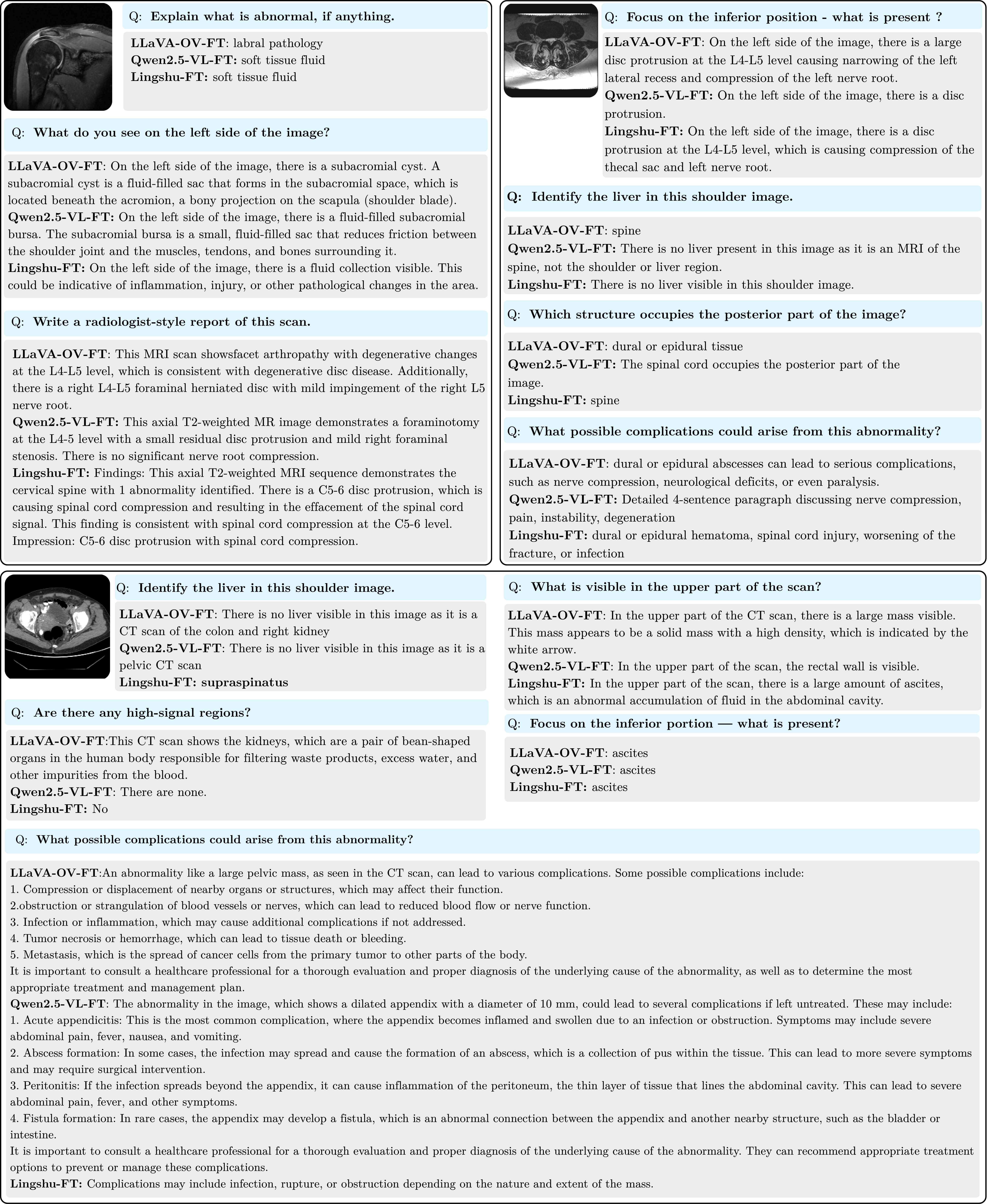}
\caption{Out-of-template evaluation of three fine-tuned VLMs (LLaVA-OneVision-FT, Qwen2.5-VL-FT, Lingshu-FT) on representative CT/MRI scans. Questions are phrased in ways not seen during training, including open-ended descriptions and higher-level reasoning queries. Cells summarize whether the model correctly identifies anatomy/pathology, provides plausible but partially incorrect answers, or hallucinates information.} %This evaluation highlights model robustness to linguistic variation, areas of reliable generalization, and remaining challenges in handling complex, clinically nuanced questions.}
\label{fig:out-of-template}
\end{figure}

Overall, the models demonstrate several encouraging behaviors. All three fine-tuned VLMs are able to correctly identify visible anatomical structures in many cases, even when questions are phrased differently from the training templates. This indicates that the models have learned meaningful visual representations and can generalize their anatomical knowledge beyond templated cues. Additionally, Qwen2.5-VL-FT and Lingshu-FT often provide clinically plausible explanations for observed abnormalities, showing that they are capable of reasoning about pathology in a way that is consistent with image content. In particular, these models handle absent structures conservatively, avoiding hallucination when organs or pathologies are not present, which reflects a careful grounding in the visual evidence. In some instances, LLaVA-OV-FT generates detailed stepwise clinical descriptions, demonstrating the potential for synthesizing information into structured, radiologist-style explanations.

The evaluation also highlights areas where improvements are needed. LLaVA-OV-FT occasionally hallucinates structures or pathologies in response to out-of-template questions, indicating reliance on linguistic patterns learned during training. Complex pathology questions, such as those involving potential complications or higher-level interpretations, are more prone to over-specification or partially incorrect reasoning, particularly for models heavily influenced by template patterns. These observations suggest that while current fine-tuned models can generalize beyond template formats, their robustness is limited when faced with more diverse or clinically nuanced question phrasing.

These findings underscore the importance of expanding the diversity of training data, both in the way questions and answers are formulated and in the structure of multi-turn interactions. By incorporating multiple paraphrases, open-ended reasoning prompts, and varied conversational styles, future work can further enhance the generalization capacity of vision-language models in radiologic VQA, reducing template dependency and improving reliability in real-world clinical settings.
}

%------------------------------------------------------------------
\subsection{Extended results on text-only evaluation}
\label{appx_subsec:text-only}

\begin{wrapfigure}{r}{0.45\textwidth}
  \vspace{-0.8cm}
  \begin{center}
    \includegraphics[width=\linewidth]{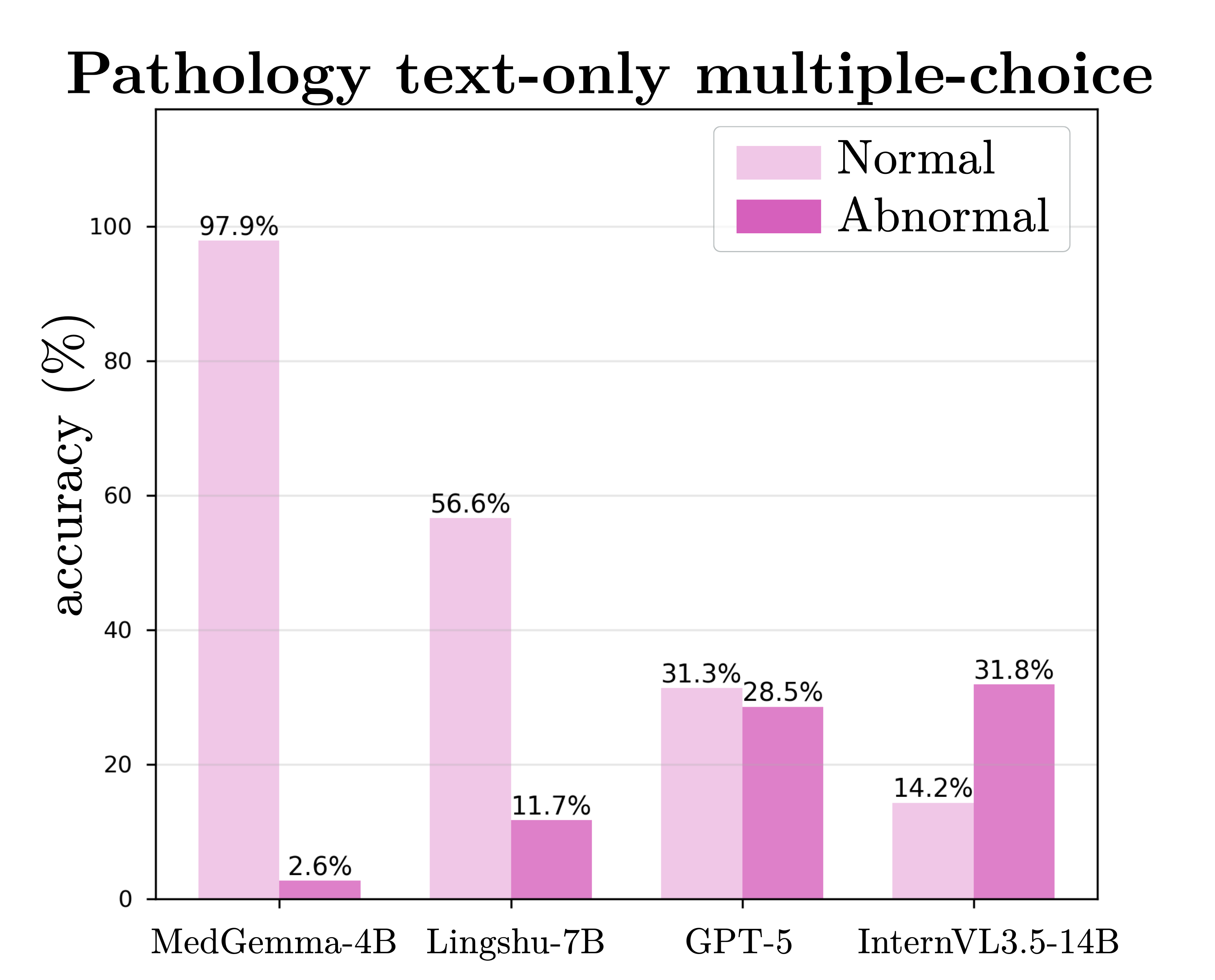}
  \end{center}
  \vspace{-0.8cm}
  \caption{MC text-only accuracy on RadImageNet-VQA.}
  %\vspace{-1.2cm}
   \label{appx_fig:text-only-mc-abn-vs-norm}
\end{wrapfigure}
Figure~\ref{fig:text-ony-open-mc} shows that text-only performance on RadImageNet-VQA drops close to the random baseline, suggesting that most multiple-choice questions cannot be solved without visual evidence. To better understand the residual accuracy that remains, Figure~\ref{appx_fig:text-only-mc-abn-vs-norm} breaks results down by abnormality status. When the image is removed, medical-oriented models such as MedGemma-4B almost always choose the “no pathology seen’’ option, which yields very high accuracy for normal cases but causes performance on abnormal cases to collapse. In contrast, general-purpose models (GPT-5 and InternVL3.5-14B) do not rely on this conservative default and instead spread predictions across all choices, resulting in lower but more balanced scores. These results show that the small amount of text-only accuracy observed on RadImageNet-VQA primarily reflects conservative guessing strategies rather than the ability to infer pathology from the question text alone, further confirming that pathology identification requires image grounding.

%------------------------------------------------------------------
\subsection{More qualitative examples}
\label{appx_subsec:examples}

\begin{figure}[H]
    \centering
    \includegraphics[width=1\linewidth]{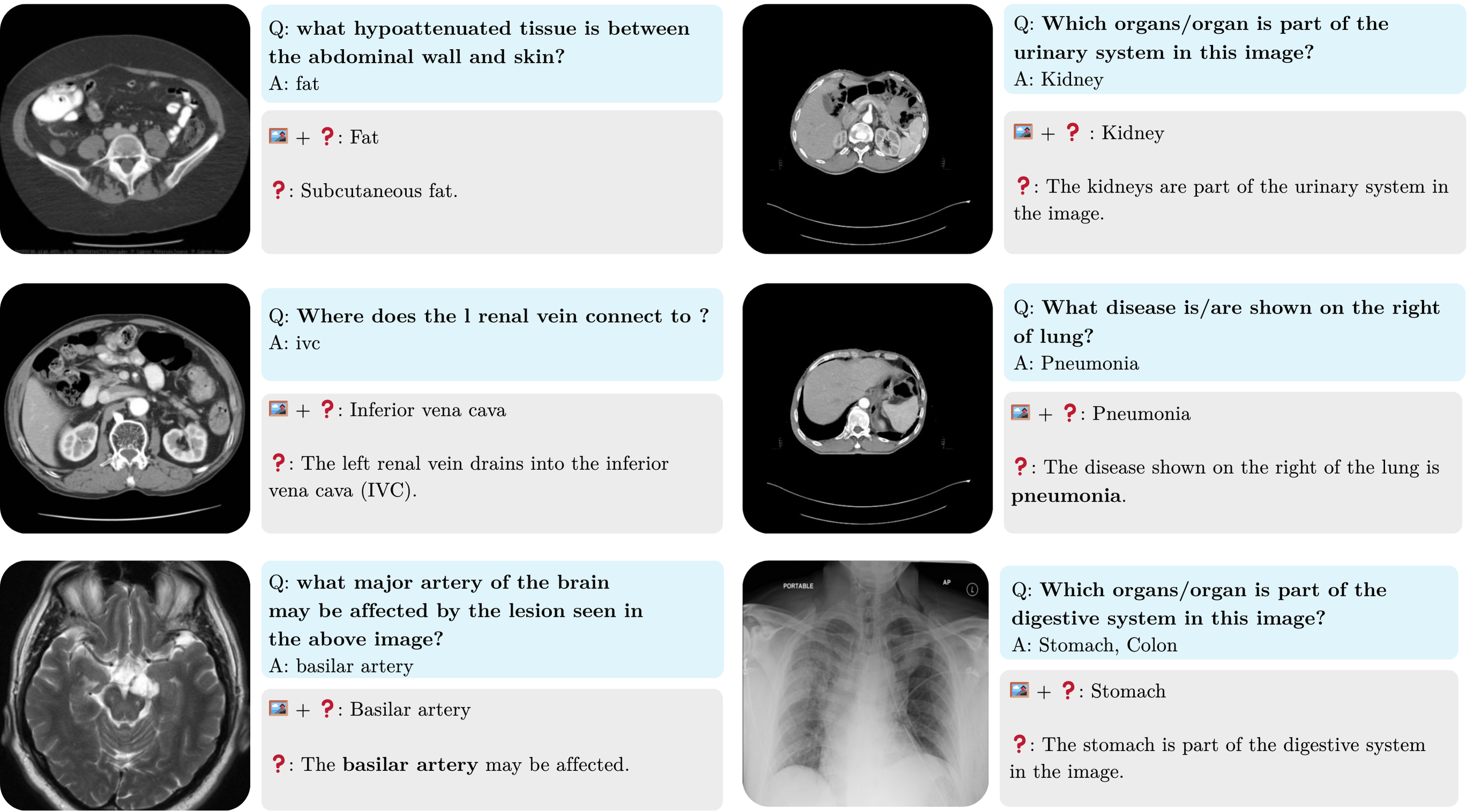}
    \caption{Examples of samples from VQA-RAD and SLAKE where current models do not need the image. The responses were obtained with MedGemma-4B.}
    \label{fig:image-vs-no-image}
\end{figure}

The examples in Figure~\ref{fig:image-vs-no-image} illustrate how VQA-RAD and SLAKE contain strong textual cues that enable models to answer correctly even without visual input. In each case, the model produces essentially the same response whether it receives the image and question or the question alone, showing that the question text often provides enough information to determine the answer. For instance, in \textit{Where does the left renal vein connect to?}, the mention of \textit{renal vein} directly implies the \textit{inferior vena cava}, a standard anatomical relationship independent of the specific CT slice. Likewise, questions such as \textit{Which organ is part of the urinary system?} or \textit{Which organs are part of the digestive system?} contain the key semantic identifiers—\textit{urinary system}, \textit{digestive system}—that uniquely determine the correct organ without requiring visual confirmation. Even pathology-oriented prompts (e.g., \textit{What disease is shown on the right lung?}) can be shortcut through frequent-label priors, leading the model to answer \textit{pneumonia} regardless of the underlying image.

\begin{figure}[H]
    \centering
    \includegraphics[width=1\linewidth]{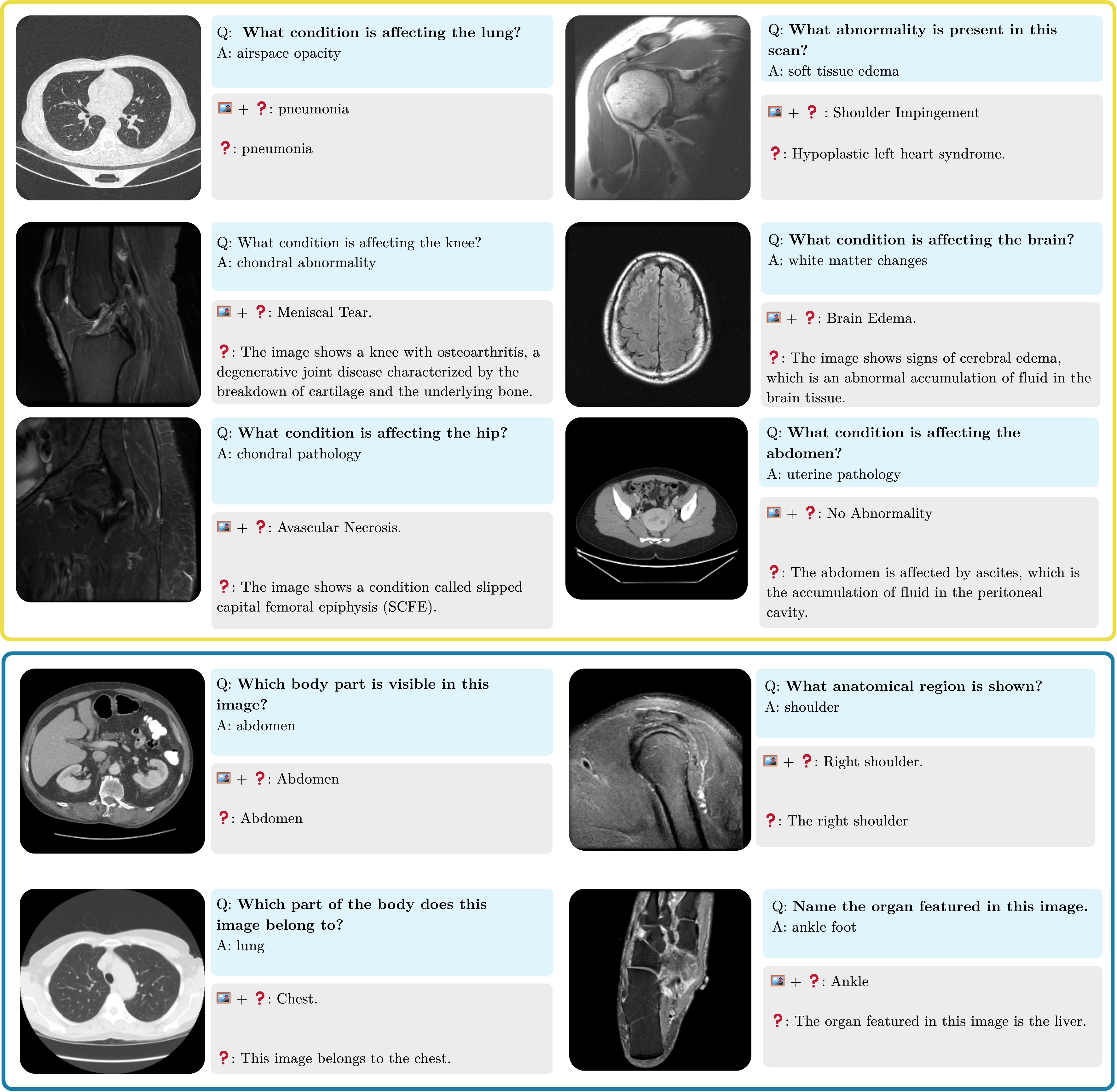}
    \caption{Examples comparing Lingshu-7B’s answers when given the image and question versus the question alone on RadImageNet-VQA.}
    \label{fig:example-lingshu-text-only-outputs}
\end{figure}

Building on the text-only results in Figure~\ref{fig:text-ony-open-mc}, where RadImageNet-VQA yields much lower shortcutability than VQA-RAD or SLAKE but where Lingshu-7B still attains 9.4\% accuracy on open-ended questions, we further inspect its predictions with and without images. Figure~\ref{fig:example-lingshu-text-only-outputs} shows qualitative examples on RadImageNet-VQA for both pathology and anatomy questions. When the prompt explicitly mentions an organ or region (e.g., lung, knee, hip, brain), the text-only model often maps this cue to a high-frequency pathology for that location—such as \textit{pneumonia} for the lung or \textit{osteoarthritis} for the knee—which sometimes coincides with the image-based answer (e.g., \textit{pneumonia}, \textit{cerebral/brain edema}) but often diverges from it (e.g., \textit{osteoarthritis} vs.\ \textit{meniscal tear, ascites} vs.\ no abnormality, \textit{SCFE} vs.\ \textit{avascular necrosis}). This behavior indicates that, in the absence of visual evidence, Lingshu falls back on learned pathology priors tied to anatomical templates rather than genuinely reasoning about the specific case. For anatomy questions, multimodal predictions are typically correct and well localized (abdomen, chest, shoulder, ankle), whereas text-only outputs tend to overuse very frequent regions such as abdomen, chest, or liver, reflecting the underlying dataset distribution more than the individual question. Overall, these examples suggest that RadImageNet-VQA still leaves a narrow band of text-based shortcuts when anatomical regions are explicitly named, but that many failures in the multimodal setting stem from the intrinsic difficulty of fine-grained pathology recognition and from the model’s reliance on organ and frequency-based priors under limited cues, rather than from purely linguistic shortcutting alone.

%==================================================================
\section{Extended Fine-Tuning Details}
\label{appx_sec:ft-details}

%------------------------------------------------------------------
\subsection{Training data composition}
\label{appx_subsec:training-compo}

To train radiologic VLMs, we assemble a multimodal corpus composed of:  
(1) RadImageNet-VQA train split. (2) additional CT/MRI converted into 2D VQA pairs  KiTS22~\cite{kits23}, and AbdomenAtlas~\cite{chen2025scaling}, where 3D volumes are sliced into 2D images and paired with captions or VQA questions derived from organ and lesion annotations;  
and (3) existing radiologic VQA datasets such as VQA-RAD~\cite{vqarad}, SLAKE~\cite{liu2021slake}, and radiology subset of LLaVA-Med~\cite{llavamed}.

\begin{table}[H]
\centering
\scriptsize
\renewcommand{\arraystretch}{1.2}
\setlength{\tabcolsep}{5pt}
\begin{tabular}{p{5cm} p{2cm} p{1.9cm} p{1.9cm}}
\toprule
\textbf{Dataset} & \textbf{\# Images} & \textbf{Alignment} & \textbf{Instruction} \\
\midrule
RadImageNet~\cite{mei2022radimagenet} & 750K & 750K & 6.75M \\
KiTS22~\cite{kits23} & 4.6K & 4.6K & 32.5K \\
AbdomenAtlas~\cite{chen2025scaling} & 43.7K & 43.7K & 211K \\
LLaVA-Med~\cite{llavamed} & 38.4K & 13.8K & 72.3K \\
VQA-RAD~\cite{vqarad} & 0.3K & -- & 1.8K \\
SLAKE~\cite{liu2021slake} & 0.4K & -- & 4.9K \\
\midrule
\textbf{Total} & \textbf{837.5K} & \textbf{812.1K} & \textbf{7.07M} \\
\bottomrule
\end{tabular}
\caption{
Summary of datasets used for alignment and instruction tuning,
including the number of images and approximate QA pair counts.
}
\label{appx_tab:datasets-collection}
\end{table}

\begin{figure}[H]
    \centering
    \includegraphics[width=1\linewidth]{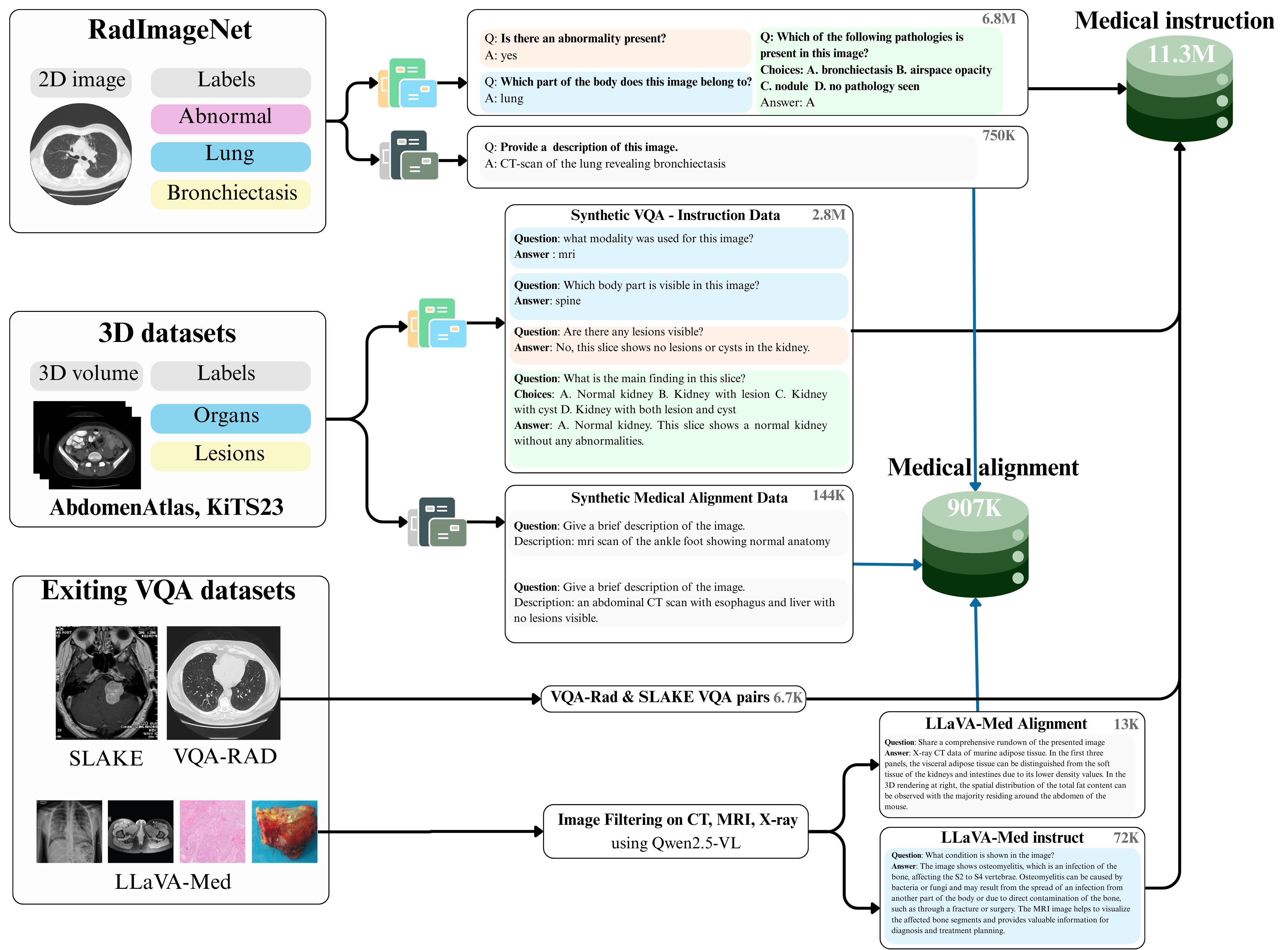}
    \vspace{-0.8cm}
    \caption{The overall data curation pipeline of radiologic multimodal data}
    \label{appx_fig:trainset-collect}
\end{figure}

Together, these three branches feed into a unified training corpus comprising medical instruction data and medical alignment data, as depicted in Figure~\ref{appx_fig:trainset-collect}.

%------------------------------------------------------------------
\subsection{Implementation Details}
\label{appx_subsec:implem-ft}

\begin{table}[H]
\centering
\footnotesize          % instead of \small
\renewcommand{\arraystretch}{1.1}   % instead of 1.25
\setlength{\tabcolsep}{4pt}         % 
\begin{tabular}{@{}p{0.32\textwidth} p{0.60\textwidth}@{}}
\toprule
\textbf{Category} & \textbf{Configuration} \\
\midrule

\textbf{Training stages} &
Visual alignment (1 epoch)\\
& Instruction tuning (2 epochs) \\

\midrule
\textbf{Batching and scheduling} &
Global batch size: 16 (4 GPUs, per-GPU batch = 1, gradient accumulation = 4) \\
& Learning rate: 1e–5 (LLM), 2e–6 (vision tower) \\
& Warm-up: 3\% of total steps \\
& LR schedule: cosine decay \\

\midrule
\textbf{Optimization} &
Optimizer: AdamW ($\beta_1$ = 0.9, $\beta_2$ = 0.999) \\
& Weight decay: 0.0 \\

\midrule
\textbf{Evaluation parameters} &
Max new tokens: 8192 \\
& Decoding: temperature = 0, top-p = 0.0001 \\
& Repetition penalty: 1.0 \\
& Judge backend: \texttt{mistral-large-latest} \\
& Random seed: 42. \\

\bottomrule
\end{tabular}
\caption{Training and evaluation configuration for RadImageNet-VQA experiments.}
\label{appx_tab:training_eval_config}
\end{table}

We fine-tune all models described in Section~\ref{sec:model-fine-tuning} with 4x \textbf{NVIDIA H100 GPUs}. Training follows a two-stage procedure: (1) a \textit{visual alignment} phase in which only the vision tower and projector are updated while the language model remains frozen, and (2) an \textit{instruction tuning} phase where all components are trainable. Evaluation uses a unified decoding configuration and a fixed LLM-as-a-judge setup for all models. The full set of hyperparameters and evaluation settings is summarized in Table~\ref{appx_tab:training_eval_config}.

%------------------------------------------------------------------
\subsection{Full fine-tuning results}

Table~\ref{appx_tab:ft_values_deltas_colored} reports the exact fine-tuned values and deltas for each task and question type on RadImageNet-VQA. This show that all models benefit substantially from radiologic instruction tuning, with very large gains on abnormality and open-ended pathology questions and near-ceiling performance on anatomy.

\label{appx_subsec:full-results-ft}

\begin{table}[!h]
\centering
\footnotesize
\renewcommand{\arraystretch}{1.2}
\begin{tabular}{l|ccc}
\toprule
\textbf{Task} 
& \textbf{LLaVA-One-Vision} 
& \textbf{Lingshu-7B} 
& \textbf{QwenVL-2.5-7B} \\
\midrule

\rowcolor{gray!15}
\multicolumn{4}{l}{\textbf{Anatomy}} \\

Open 
& 99.4 \textcolor{ForestGreen}{$\uparrow$(+51.0)}
& 99.4 \textcolor{ForestGreen}{$\uparrow$(+49.8)}
& 99.3 \textcolor{ForestGreen}{$\uparrow$(+61.8)}
\\

Closed+ 
& 87.4 \textcolor{ForestGreen}{$\uparrow$(+4.7)}
& 95.2 \textcolor{ForestGreen}{$\uparrow$(+4.5)}
& 93.4 \textcolor{ForestGreen}{$\uparrow$(+8.5)}
\\

Closed– 
& 97.8 \textcolor{ForestGreen}{$\uparrow$(+16.5)}
& 97.6 \textcolor{ForestGreen}{$\uparrow$(+12.5)}
& 97.3 \textcolor{ForestGreen}{$\uparrow$(+18.2)}
\\

MC 
& 99.4 \textcolor{ForestGreen}{$\uparrow$(+10.7)}
& 98.9 \textcolor{ForestGreen}{$\uparrow$(+10.0)}
& 98.5 \textcolor{ForestGreen}{$\uparrow$(+18.0)}
\\

\midrule
\rowcolor{gray!15}
\multicolumn{4}{l}{\textbf{Abnormality}} \\

Closed
& 87.6 \textcolor{ForestGreen}{$\uparrow$(+37.8)}
& 87.7 \textcolor{ForestGreen}{$\uparrow$(+39.8)}
& 85.9 \textcolor{ForestGreen}{$\uparrow$(+16.4)}
\\

\midrule
\rowcolor{gray!15}
\multicolumn{4}{l}{\textbf{Pathology}} \\

Open 
& 42.6 \textcolor{ForestGreen}{$\uparrow$(+26.6)}
& 41.2 \textcolor{ForestGreen}{$\uparrow$(+25.5)}
& 39.2 \textcolor{ForestGreen}{$\uparrow$(+29.4)}
\\

Closed+ 
& 67.2 \textcolor{ForestGreen}{$\uparrow$(+11.9)}
& 84.3 \textcolor{ForestGreen}{$\uparrow$(+27.3)}
& 74.7 \textcolor{ForestGreen}{$\uparrow$(+5.5)}
\\

Closed– 
& 77.2 \textcolor{ForestGreen}{$\uparrow$(+15.9)}
& 65.3 \textcolor{red}{$\downarrow$(-13.5)}
& 68.1 \textcolor{ForestGreen}{$\uparrow$(+20.7)}
\\

MC 
& 60.1 \textcolor{ForestGreen}{$\uparrow$(+26.5)}
& 49.9 \textcolor{ForestGreen}{$\uparrow$(+20.3)}
& 53.4 \textcolor{ForestGreen}{$\uparrow$(+23.3)}
\\

\midrule
\textbf{Avg} 
& 79.9 \textcolor{ForestGreen}{$\uparrow$(+22.4)}
& 79.9 \textcolor{ForestGreen}{$\uparrow$(+19.5)}
& 78.9 \textcolor{ForestGreen}{$\uparrow$(+22.5)}
\\

\bottomrule
\end{tabular}
\caption{
Fine-tuned performance on RadImageNet-VQA with delta values relative to zero-shot baselines.
Improvements are shown in \textcolor{ForestGreen}{green} and regressions in \textcolor{red}{red}.
}
\label{appx_tab:ft_values_deltas_colored}
\end{table}

\end{document}